\documentclass[acmtog]{acmart}
\usepackage{booktabs} 
\usepackage{comment}

\newcommand{\Fref}[1]{Figure~\ref{#1}}
\newcommand{\Sref}[1]{Section~\ref{#1}}

\newcommand{\eref}[1]{Eq.~(\ref{#1})}
\newcommand{\fref}[1]{Fig.~\ref{#1}}

\usepackage[british,UKenglish,USenglish,english,american]{babel}

\newcommand{\eg}{\textit{e.g.}}
\newcommand{\ie}{\textit{i.e.}}

\setcopyright{none}
\acmJournal{TOG}
\acmYear{2018}\acmVolume{37}\acmNumber{6}\acmArticle{244}\acmMonth{11} \acmDOI{10.1145/3272127.3275046}

\setcitestyle{square}

\begin{document}

\title{CariGANs: Unpaired Photo-to-Caricature Translation}

\author{Kaidi Cao}
\authornote{Project page: \url{https://cari-gan.github.io/}}
\authornote{This work was done when Kaidi Cao was an intern at Microsoft Research Asia.}
\affiliation{%
  \department{Department of Electronic Engineering}
  \institution{Tsinghua University}}

\author{Jing Liao}
\authornote{indicates corresponding author.}
\affiliation{%
  \institution{City University of Hong Kong,}
  \institution{Microsoft Research}}

\author{Lu Yuan}
\affiliation{%
  \institution{Microsoft AI Perception and Mixed Reality}}

\renewcommand{\shortauthors}{Cao, Liao and Yuan}

\begin{abstract}Facial caricature is an art form of drawing faces in an exaggerated way to convey humor or sarcasm. In this paper, we propose the first Generative Adversarial Network (GAN) for unpaired photo-to-caricature translation, which we call ``CariGANs". It explicitly models geometric exaggeration and appearance stylization using two components: \emph{CariGeoGAN}, which only models the geometry-to-geometry transformation from face photos to caricatures, and \emph{CariStyGAN}, which transfers the style appearance from caricatures to face photos without any geometry deformation. In this way, a difficult cross-domain translation problem is decoupled into two easier tasks. The perceptual study shows that caricatures generated by our \emph{CariGANs} are closer to the hand-drawn ones, and at the same time better persevere the identity, compared to state-of-the-art methods. Moreover, our \emph{CariGANs} allow users to control the shape exaggeration degree and change the color/texture style by tuning the parameters or giving an example caricature.

\end{abstract}



\begin{teaserfigure}
\centering
\setlength{\tabcolsep}{0.003\linewidth}
  \scalebox{1.0}{
\begin{tabular}{cccccc}
\includegraphics[height=0.155\linewidth]
			{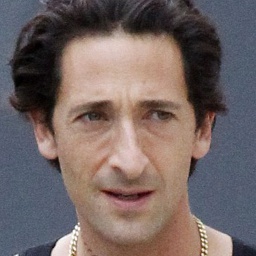}&
            \includegraphics[height=0.155\linewidth]
			{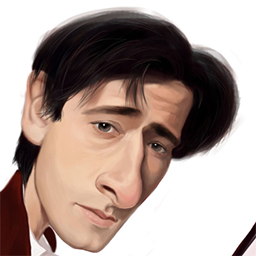}&
            \includegraphics[height=0.155\linewidth]
			{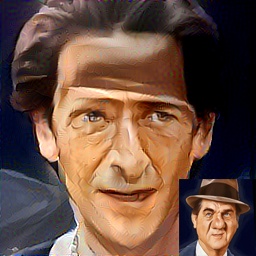}&
            \includegraphics[height=0.155\linewidth]
			{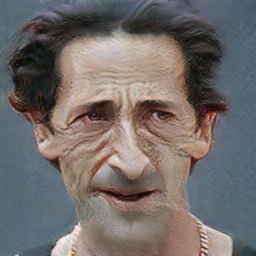}&
             \includegraphics[height=0.155\linewidth]
			{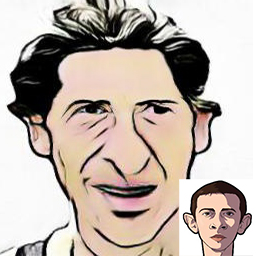}&
             \includegraphics[height=0.155\linewidth]
			{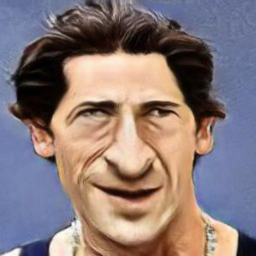}\\
            \includegraphics[height=0.155\linewidth]
			{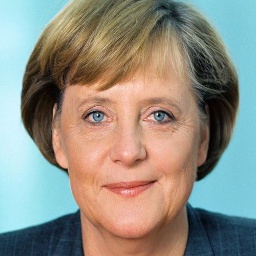}&
            \includegraphics[height=0.155\linewidth]
			{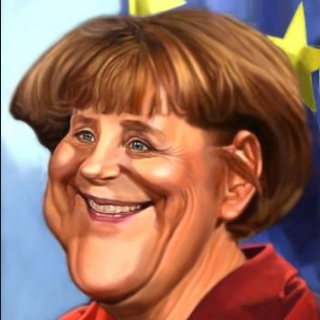}&
            \includegraphics[height=0.155\linewidth]
			{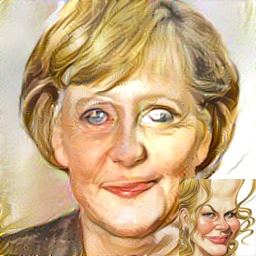}&
            \includegraphics[height=0.155\linewidth]
			{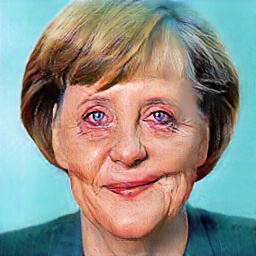}&
             \includegraphics[height=0.155\linewidth]
			{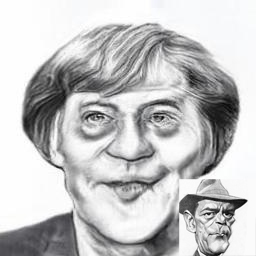}&
             \includegraphics[height=0.155\linewidth]
			{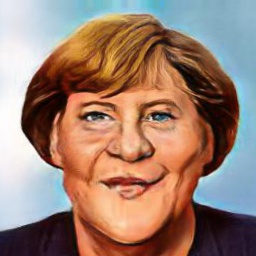}\\
(a) Photo    & (b) Hand-drawn   & (c) \citet{gatys2015neural} & (d) \citet{CycleGAN2017} & (e)  Ours (with ref)  &  (f) Ours (with noise) 
\end{tabular}}
\vspace*{-0.1in}
\caption{Comparison of the caricature drawn manually (b) and generated automatically with neural style transfer \cite{gatys2015neural} (c), CycleGan \cite{CycleGAN2017} (d), and Our CariGANs with a given reference (e) or a random noise (f). Please note networks used in (d)(e)(f) are trained with the same dataset. And the reference used in the result is overlaid on its bottom-right corner. Photos: MS-Celeb-1M dataset, hand-drawn caricatures (from top to bottom): \textcopyright Lucy Feng/deviantart, \textcopyright Tonio/toonpool.}
\label{fig:intro}
\end{teaserfigure}

\maketitle

\section{Introduction}

A caricature can be defined as an art form of drawing persons (usually faces) in a simplified or exaggerated way through sketching, pencil strokes, or other artistic drawings. As a way to convey humor or sarcasm, caricatures are commonly used in entertainment, and as gifts or souvenirs, often drawn by street vendors. Artists have the amazing ability to capture distinct facial features of the subject from others, and then exaggerate those features.

There have been a few attempts to interactively synthesize facial caricature~\cite{akleman1997making,akleman2000making,chen2002pictoon,gooch2004human}, but it requires professional skills to produce expressive results. A few automatic systems are proposed, which rely on hand-crafted rules ~\cite{brennan2007caricature,koshimizu1999kansei,mo2004improved,liang2002example}, often derived from the drawing procedure of artists. However, these approaches are restricted to a particular artistic style, \eg, sketch or a certain cartoon, and predefined templates of exaggeration.

In recent years, deep learning, as the representative technique of learning from examples (especially from big data), has been successfully used in image-to-image translation \cite{hinton2006reducing,isola2017image,yi2017dualgan, kim2017learning,zhu2017toward,liu2017unsupervised,huang2018munit}. As is commonly known, most photo and caricature examples are unfortunately \emph{unpaired} in the world. So the translation may be infeasible to be trained in a supervised way like Autoencoder~\cite{hinton2006reducing}, Pix2Pix~\cite{isola2017image}, and other paired image translation networks. Building such a dataset with thousands of image pairs  (\ie, a face photo and its associate caricature drawn by artists) would be too expensive and tedious.

On the other hand, there are two keys to generating a caricature: shape exaggeration and appearance stylization, as shown in \fref{fig:intro} (a)(b). Neural style transfer methods~\cite{gatys2015neural,liao2017visual,johnson2016perceptual}, which transfer the artistic style from a given reference to a photo through deep neural networks, are good at stylizing appearances, but do not exaggerate the geometry, as shown in \fref{fig:intro} (c). There are a few works ~\cite{liu2017unsupervised,CycleGAN2017,zhu2017toward,huang2018munit} proposed for unsupervised cross-domain image translation, which in principle will learn both geometric deformation and appearance translation simultaneously. However, the large gap of shape and appearance between photos and caricatures imposes a big challenge to these networks, and thus they generate unpleasant results, as shown in \fref{fig:intro} (d).

In order to generate a reasonable result approaching caricature artists' productions, one has to ask ``what is the desired quality of caricature generation?". Shape exaggeration is not a distortion, which is complete denial of truth~\cite{redman1984draw}. The exaggerated shape should maintain the relative geometric location of facial components, and only emphasize the subject's features, distinct from others. The final appearance should be faithful to visual styles of caricatures, and keep the \emph{identity} with the input face, as addressed in other face generators~\cite{brennan2007caricature,mo2004improved,liang2002example}. Moreover, the generation must be diverse and controllable. Given one input face photo, it allows for the generation of variant types of caricatures, and even controls the results either by example caricature, or by user interactions (\eg, tweaking exaggerated shape). It can be useful and complementary to existing interactive caricature systems.

In this paper, we propose the first Generative Adversarial Network (GAN) for unpaired photo-to-caricature translation, which we call ``CariGANs". It explicitly models geometric exaggeration and appearance stylization using two components: \emph{CariGeoGAN}, which only models the geometry-to-geometry transformation from face photos to caricatures, and \emph{CariStyGAN}, which transfers the style appearance from caricatures to face photos without any geometry deformation. Two GANs are separately trained for each task, which makes the learning more robust. To build the relation between unpaired image pairs, both \emph{CariGeoGAN} and \emph{CariStyGAN} use cycle-consistency network structures, which are widely used in cross-domain or unsupervised image translation ~\cite{zhu2017toward, huang2018munit}. Finally, the exaggerated shape (obtained from \emph{CariGeoGAN}) serves to exaggerate the stylized face (obtained from \emph{CariStyGAN}) via image warping.

In \emph{CariGeoGAN}, we use the PCA representation of facial landmarks instead of landmarks themselves as the input and output of GAN. This representation implicitly enforces the constraint of face shape prior in the network. Besides, we consider a new characteristic loss in \emph{CariGeoGAN} to encourage exaggerations of distinct facial features only, and avoid arbitrary distortions. Our \emph{CariGeoGAN} outputs the landmark positions instead of the image, so the exaggeration degree can be tweaked before the image warping. It makes results controllable and diverse in geometry.

As to the stylization, our \emph{CariStyGAN} is designed for pixel-to-pixel style transfer without any geometric deformation. To exclude the interference of geometry in training \emph{CariStyGAN}, we create an intermediate caricature dataset by warping all original caricatures to the shapes of photos via the reverse geometry mapping derived from \emph{CariGeoGAN}. In this way, the geometry-to-geometry translation achieved by \emph{CariGeoGAN} is successfully decoupled from the appearance-to-appearance translation achieved by \emph{CariStyGAN}. In addition, our \emph{CariStyGAN} allows multi-modal image translation, which traverses the caricature style space by varying the input noise. It also supports example-guided image translation, in which the style of the translation outputs are controlled by a user-provided example caricature. To further keep identity in appearance stylization, we add perceptual loss \citep{johnson2016perceptual} into \emph{CariStyGAN}. It constrains the stylized result to preserve the content information of the input.

With our \emph{CariGAN}, the photos of faces in the wild can be automatically translated to caricatures with geometric exaggeration and appearance stylization, as shown in \fref{fig:intro} (f). We have extensively compared our method with state-of-the-art approaches. The perceptual study results show caricatures generated by our \emph{CariGANs} are closer to the hand-drawn caricatures, and at the same time better persevere the identity, compared to the state-of-the-art. We further extend the approach to new applications, including generating video caricatures, and converting a caricature to a real face photo.

In summary, our key contributions are:
\begin{enumerate}
	\item We present the first deep neural network for unpaired photo-to-caricature translation. It achieves both geometric exaggeration and appearance stylization by explicitly modeling the translation of geometry and appearance with two separate GANs.
	\item We present \emph{CariGeoGAN} for geometry exaggeration, which is the first attempt to use cycle-consisteny GAN for cross-domain translation in geometry. To constrain the shape exaggeration, we adopt two major novel extensions, like PCA representation of landmarks, and a characteristic loss.
	\item We present \emph{CariStyGAN} for appearance stylization, which allows multi-modal image translation, while preserving the identity in the generated caricature by adding a perceptual loss.
	\item Our \emph{CariGANs} allows user to control the exaggeration degree in geometric and appearance style by simply tuning the parameters or giving an example caricature.
\end{enumerate}

\section{Related Work}

Recent literature suggests two main directions to tackle the photo-to-caricature transfer task: traditional graphics-based methods and recent deep learning-based methods.

\paragraph{\textbf{Graphics-based methods.}}
In computer graphics, translating photo to caricature or cartoon is interesting, and has been studied for a long while. These techniques can be categorized into three groups.

The category develops deformation systems which allow users to manipulate photos interactively~\cite{akleman1997making,akleman2000making,chen2002pictoon,gooch2004human}. These kind of methods usually require expert knowledge and detailed involvement of experienced artists.

The second category defines hand-craft rules to automatically exaggerate difference from the mean (EDFM). Brennan~\cite{brennan2007caricature} is the first to present the EDFM idea. Some following works~\cite{koshimizu1999kansei,mo2004improved,tseng2007synthesis,le2011shape,liu2006mapping,liao2004automatic} improve rules of EDFM to represent the distinctiveness of the facial features better. Besides 2D exaggeration, there is also some work utilizing tensor-based 3D model to exaggerate facial features ~\cite{yang2012facial}. However there is a central question regarding the effectiveness of EDFM: whether these hand-crafted rules faithfully reflect the drawing styles of caricaturists. 

\begin{figure*}
\centering
\includegraphics[width=0.95\textwidth]{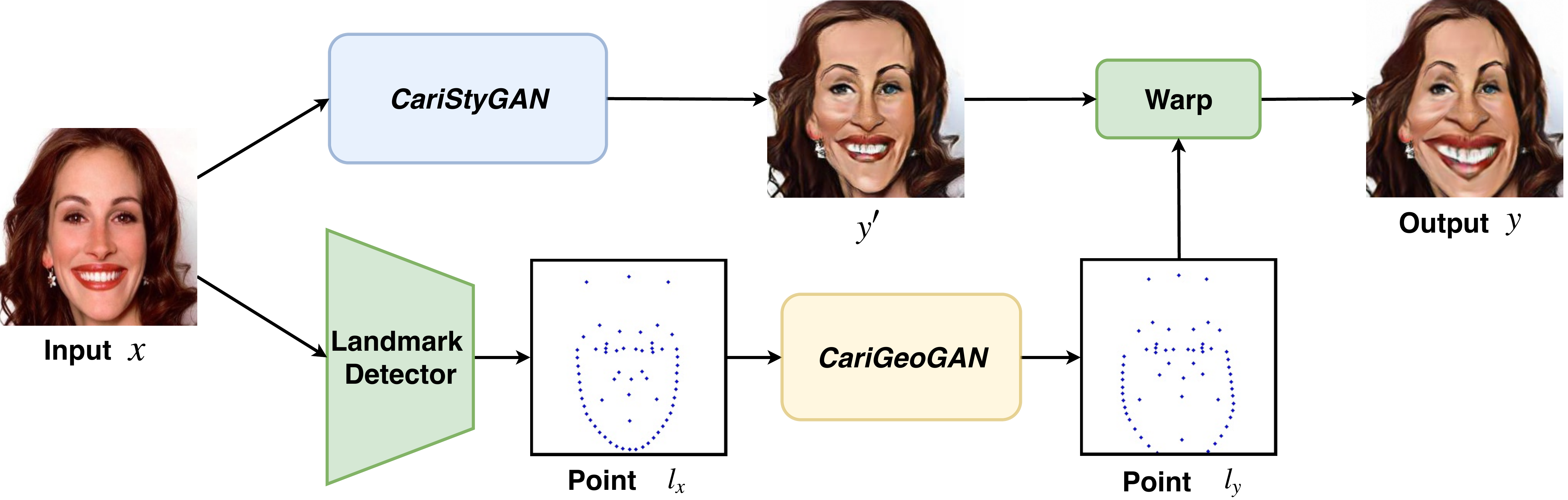}
\vspace*{-0.1in}
\caption{Overall Pipeline of Proposed Method. Input image: CelebA dataset.}
\label{fig:Pipeline}

\end{figure*}

The third category of methods directly learn rules from paired photo-caricature images, drawn by caricaturists. For examples, Liang et al.\cite{liang2002example} propose learning prototypes by analyzing the correlation between the image caricature pairs using
partial least-squares (PLS). Shet et al. \cite{shet2005use} train a Cascade Correlation Neural Network (CCNN) network to capture the drawing style in relation to facial components. In practice, however, it is difficult to obtain a large paired training set. Learning from one-shot or a few exemplars makes it ineffective to cover the variances of existing caricatures.

\paragraph{\textbf{Neural style transfer.}}
Recently, inspired by the power of CNN, the pioneering work of Gatys et al. \cite{gatys2015neural} presents a general solution to transfer
the style of a given artwork to any image automatically. Many follow-up works have been proposed to improve quality \cite{sziranyi1997markov,liao2017visual}, speed \cite{johnson2016perceptual,chen2017stylebank}, or video extension \cite{chen2017coherent}. Notwithstanding their success in transferring photos or videos into many artistic styles like pencil, watercolor, oil painting and etc., they fail to generate caricatures with geometry exaggerations since these methods transfer textures and colors of a specific style while preserving the image content.

\paragraph{\textbf{Image-to-image translation networks.}}
There are a series of work based on the GAN proposed for a general image-to-image translation. Isola et al. \cite{isola2017image} develop the pix2pix network trained with the supervision of images pairs and achieve reasonable results on many translation tasks such as photo-to-label, photo-to-sketch and photo-to-map. BicycleGAN~\cite{zhu2017toward} extends it to multi-modal translation. Some networks including CycleGAN~\cite{CycleGAN2017}, DualGAN~\cite{yi2017dualgan}, DiscoGAN~\cite{kim2017learning}, UNIT~\cite{liu2017unsupervised}, DTN~\cite{taigman2016unsupervised} etc. have been proposed for unpaired one-to-one translation, while MNUIT~\cite{huang2018munit} was proposed for unpaired many-to-many translation. These networks often succeed on the unpaired translation tasks which are restricted to color or texture changes only, \eg, horse to zebra, summer to winter. For photo-to-caricature translation, they fail to model both geometric and appearance changes. By contrast, we explicitly model the two translations by two separated GANs: one for geometry-to-geometry mapping and another for appearance-to-appearance translation. Both GANs respectively adopt the cycle-consistent network structures (\eg, CycleGAN~\cite{CycleGAN2017}, MNUIT~\cite{huang2018munit}) since each type of translation still builds on unpaired training images.

\section{Method}

For caricature generation, previous methods, based on learning from examples, rely on paired photo-to-caricature images. Artists are required to paint corresponding caricatures for each photo. So it is infeasible to build such a paired image dataset for supervised learning due to high cost in money and time. In fact, there are a great number of caricature images found in the Internet, \eg, Pinterest.com. How to learn the photo-to-caricature translation from unpaired photos and caricatures is our goal. Meanwhile, the generated caricature should preserve the identity of the face photo.

Let $X$ and $Y$ be the face photo domain and the caricature domain respectively, where no pairing exists between the two domains. For the photo domain $X$, we randomly sample $10,000$ face images from the CelebA database~\cite{liu2015faceattributes} $\{x_i\}_{i={1,...,N}}, x_i \in X$ which covers diverse gender, races, ages, expressions, poses and etc. To obtain the caricature domain $Y$, we collect $8,451$ hand-drawn portrait caricatures from the Internet with different drawing styles (\eg, cartoon, pencil-drawing) and various exaggerated facial features, $\{y_i\}_{i={1,...,M}}, y_i \in Y$. We want to learn a mapping $\Phi: X  \rightarrow Y$ that can transfer an input $x \in X$ to a sample $y=\Phi(x)$, $y \in Y$. This is a typical problem of cross-domain image translation, since photo domain and caricature domain may be obviously different in both geometry shape and texture appearance. We cannot directly learn the mapping form $X$ to $Y$ by other existing image-to-image translation networks. Instead, we decouple $\Phi$ into two mappings $\Phi_{geo}$ and $\Phi_{app}$ for geometry and appearance respectively.

\fref{fig:Pipeline} illustrates our two-stage framework, where two mappings $\Phi_{geo}$ and $\Phi_{app}$ are respectively learnt by two GANs. In the first stage, we use \emph{CariGeoGAN} to learn geometry-to-geometry translation from photo to caricature. Geometric information is represented with facial landmarks. Let $L_X$, $L_Y$ be the domains of face landmarks (from $X$) and caricature landmarks (from $Y$) respectively. In inference, the face landmarks $l_x$ of the face photo $x$ can be automatically estimated from an existing face landmark detector module. Then, \emph{CariGeoGAN} learns the mapping $\Phi_{geo}: L_X \rightarrow L_Y$ to exaggerate the shape, generating the caricature landmark $l_y \in L_Y$. In the second stage, we use \emph{CariStyGAN} to learn the appearance-to-appearance translation from photo to caricature while preserving its geometry. Here, we need to synthesize an intermediate result $y' \in Y'$, which is assumed to be as close as caricature domain $Y$ in appearance and as similar as photo domain $X$ in shape. The appearance mapping is denoted as $\Phi_{app} : X \rightarrow Y'$. Finally, we get the final output caricature $y \in Y$ by warping the intermediate stylization result $y'$ with the guidance of exaggerated landmarks $l_y$. The warping is done by a differentiable spline interpolation module \cite{cole2017synthesizing}.

In next sections, we will describe the two GANs in detail.

\subsection{Geometry Exaggeration}

In this section, we present \emph{CariGeoGAN} which learns geometric exaggeration of the distinctive facial features.

\begin{figure}[t]
	\footnotesize
	\setlength{\tabcolsep}{0.003\linewidth}
    \scalebox{1.1}{
		\begin{tabular}{cccc}
            \includegraphics[width=0.22\linewidth]
			{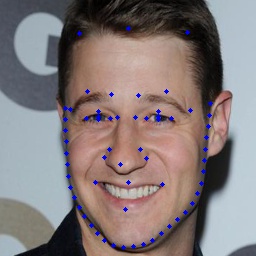}&
            \includegraphics[width=0.22\linewidth]
			{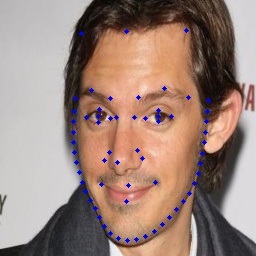}&
            \includegraphics[width=0.22\linewidth]
			{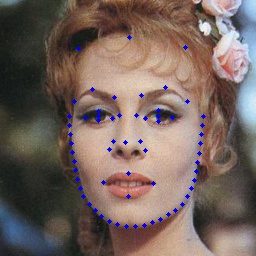}&
            \includegraphics[width=0.22\linewidth]
			{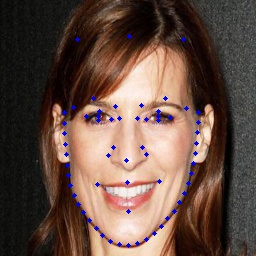} \\
            \includegraphics[width=0.22\linewidth]
			{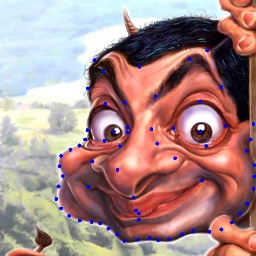}&
            \includegraphics[width=0.22\linewidth]
			{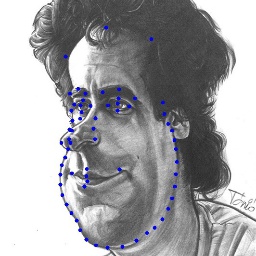}&
            \includegraphics[width=0.22\linewidth]
			{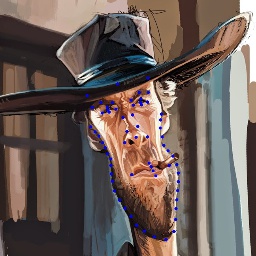}&
            \includegraphics[width=0.22\linewidth]
			{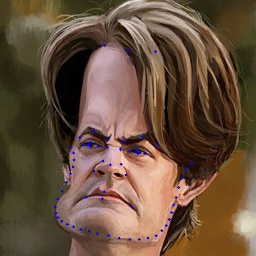}
		\end{tabular}}
\vspace*{-0.1in}
    \caption {Some samples from database of portrait photos (upper row) and caricatures (lower row). Photos: CelebA dataset, caricatures (from left to right): \textcopyright Tonio/toonpool, \textcopyright Tonio/toonpool, \textcopyright Alberto Russo/www.dessins.ch, \textcopyright Alberto Russo/www.dessins.ch.}
	\label{fig:sampledata}
\end{figure}

\paragraph{\textbf{Training data}}
Face shape can be represented by 2D face landmarks either for real face photos $X$, or for caricatures $Y$. We manually label $63$ facial landmarks for each image in both $X$ and $Y$. For the annotation, we show overlaid landmarks in \fref{fig:sampledata}. To centralize all facial shapes, all images in both $X$ and $Y$ are aligned to the mean face via three landmarks (center of two eyes and center of the mouth) using affine transformation. In addition, all images are cropped to the face region, and resized to $256 \times 256$ resolution for normalizing the scale. \fref{fig:sampledata} shows several transformed images with overlaid landmarks. Note that the dataset of real faces is also used to finetune a landmark detector (\citep{zhu2016unconstrained}), which supports automatic landmark detection in the inference stage.

For our \emph{CariGeoGAN}, to further reduce dimensions of its input and output, we further apply principal component analysis (PCA) on the landmarks of all samples in
$X$ and $Y$. We take the top $32$ principal components to recover $99.03\%$ of total variants. Then the $63$ landmarks of each sample are represented by a vector of $32$ PCA coefficients. This representation helps constrain the face structure during mapping learning. We will discuss its role in \Sref{sec:pca}. Let $L_X, L_Y$ be the PCA landmark domains of $X$ and $Y$, respectively. Our \emph{CariGeoGAN} learns the translation from $L_X$ to $L_Y$ instead.

\begin{figure*}[t]
\centering
\includegraphics[width=0.95\textwidth]{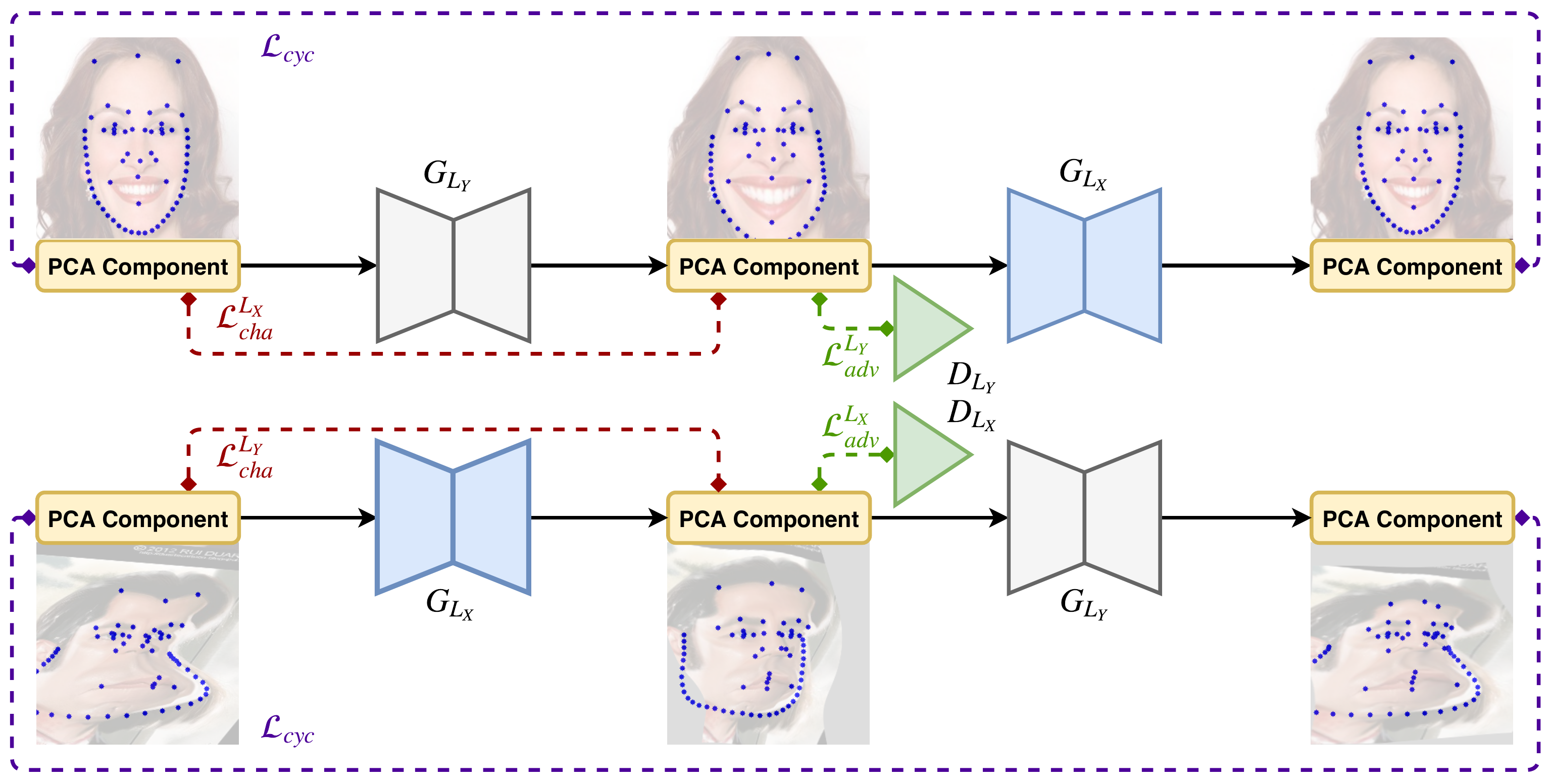}
\vspace{-0.1in}
\caption{Architecture of \emph{CariGeoGAN}. It basically follows the network structure of CycleGAN with cycle Loss $\mathcal{L}_{cyc}$ and adversarial loss $\mathcal{L}_{gan}$. But our input and output are vectors instead of images, and we add a characteristic loss $\mathcal{L}_{cha}$ to exaggerate the subject's distinct features. Input images: CelebA dataset.}
\label{fig:ImageMorphing2}
\end{figure*}

\paragraph{\textbf{CariGeoGAN}}
Since samples in $L_X$ and $L_Y$ are unpaired, the mapping function $\Phi_{geo}: L_X\rightarrow L_Y$ is highly under-constrained. CycleGAN~\cite{CycleGAN2017} couples it with a reverse mapping $\Phi^{-1}_{geo}: L_Y\rightarrow L_X$. This idea has been successfully applied for unpaired image-to-image translation, \eg, texture/color transfer. Our \emph{CariGeoGAN} is inspired by the network architecture of CycleGAN. It contains two generators and two discriminators as shown in \fref{fig:ImageMorphing2}. The forward generator $G_{L_Y}$ learns the mapping $\Phi_{geo}$ and synthesizes caricature shape $\widehat{l_y}$; while the backward generator $G_{L_X}$ learns the reverse mapping $\Phi^{-1}_{geo}$ and synthesizes face shape $\widehat{l_x}$. The discriminator $D_{L_X}$ (or $D_{L_Y}$) learn to distinguish real samples from $L_X$ (or $L_Y)$ and synthesized sample $\widehat{l_x}$ (or $\widehat{l_y}$).

The architecture of \emph{CariGeoGAN} consists of two paths. One path models the mapping $\Phi_{geo}$, shown in the top row of \fref{fig:ImageMorphing2}. Given a face shape $l_x \in L_x$, we can synthesize a caricature shape $\widehat{l_y} = G_{L_Y}(l_x)$. On one hand, $\widehat{l_y}$ is fed to the discriminator $D_{L_Y}$. On the other hand, $\widehat{l_y}$ can get back to approximate the input shape through the generator $G_{L_X}$. Similar operations are applied to the other path, which models the reverse mapping $\Phi^{-1}_{geo}$, shown in the bottom row of \fref{fig:ImageMorphing2}. Note that $G_{L_X}$ (or $G_{L_Y}$) shares weights in both pathes.

Our \emph{CariGeoGAN} is different from CycleGAN since it takes PCA vector instead of image as input and output. To incorporate the PCA landmark representation with GAN, we replace all CONV-ReLu blocks with FC-ReLu blocks in both generators and discriminators.

\paragraph{\textbf{Loss}}
We define three types of loss in the \emph{CariGeoGAN}, which are shown in \fref{fig:ImageMorphing2}.

The first is the adversarial loss, which is widely used in GANs. Specifically, we adopt the adversarial loss of LSGAN~\cite{mao2017least} to encourage generating landmarks indistinguishable from the hand-drawn caricature landmarks sampled from domain $L_Y$:

\begin{align}
\mathcal{L}_{\text{adv}}^{L_Y}(G_{L_Y},D_{L_Y}) =&  \mathbb{E}_{l_y \sim L_Y}[( D_{L_Y}(l_y) - 1)^2] \nonumber \\
             +&\mathbb{E}_{l_x \sim L_X}[D_{L_Y}(G_{L_Y}(l_x)^2].
\label{eq:adv}
\end{align}

Symmetrically, we also apply adversarial loss to encourage $G_{L_Y}$ to generate portrait photo landmarks that cannot be distinguished by the adversary $D_{L_X}$. The loss $\mathcal{L}_{\text{adv}}^{L_X}(G_{L_X},D_{L_X})$ is similarly defined as \eref{eq:adv}.

The second is the bidirectional cycle-consistency loss, which is also used in CycleGAN to constrain the cycle consistency between the forward mapping $\Phi_{geo}$ and the backward mapping $\Phi^{-1}_{geo}$. The idea is that if we apply exaggeration to $l_x$ with $G_{L_Y}$, we should get back to the input $l_x$ exactly with $G_{L_X}$, \ie, $G_{L_X}(G_{L_Y}(l_x)) \simeq l_x$. The consistency in the reverse direction $G_{L_Y}(G_{L_X}(l_y)) \simeq l_y$ is defined similarly. The loss is defined as:
\begin{align}
\mathcal{L}_{\text{cyc}}(G_{L_Y}, G_{L_X}) =  &\mathbb{E}_{l_x\sim L_X}[||G_{L_X}(G_{L_Y}(l_x))-l_x||_1] \nonumber \\
+ &\mathbb{E}_{l_y\sim L_Y}[||G_{L_Y}(G_{L_X}(l_y))-l_y||_1].
\end{align}

Cycle-consistency loss further helps constrain the mapping solution from the input to the output. However, it is still weak to guarantee that the predicted deformation can capture the distinct facial features and then exaggerate them. The third is a new characteristic loss, which penalizes the cosine differences between input landmark $l_x \in L_X$ and the predicted one $G_{L_Y}(l_x)$ after subtracting its corresponding means:
\begin{align}
  \mathcal{L}_{\text{cha}}^{L_Y}(G_{L_Y}) = &\mathbb{E}_{l_x\sim L_X}[1-\cos(l_x - \overline{L_X}, G_{L_Y}(l_x) - \overline{L_Y})],
\end{align}
where $\overline{L_X}$ (or $\overline{L_Y})$ denotes the averages of $L_X$ (or $L_Y)$. The characteristic loss in the reverse direction $\mathcal{L}_{\text{cha}}^{L_X}(G_{L_X})$ is defined similarly. The underlying idea is that the differences from a face to the mean face represent its most distinctive features and thus should be kept after exaggeration. For example, if a face has a larger nose compared to a normal face, this distinctiveness will be preserved or even exaggerated after converting to caricature.

\begin{figure}
	\footnotesize
	\setlength{\tabcolsep}{0.003\linewidth}
    \scalebox{1.1}{
		\begin{tabular}{cccc}
			\includegraphics[height=0.2\linewidth]
			{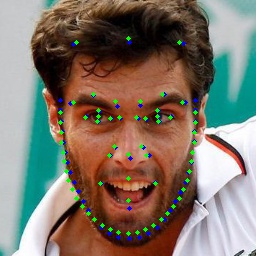}&
            \includegraphics[height=0.2\linewidth]
			{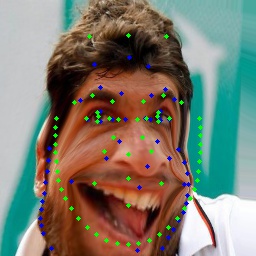}&
            \includegraphics[height=0.2\linewidth]
			{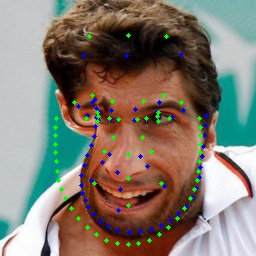}&
            \includegraphics[height=0.2\linewidth]
			{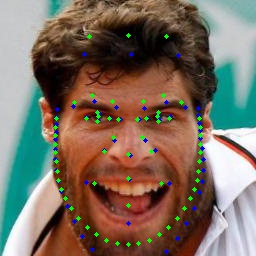}
            \\

            \includegraphics[height=0.2\linewidth]
			{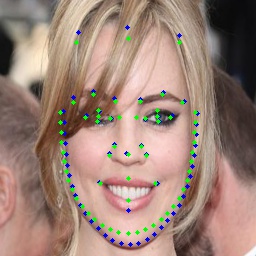}&
            \includegraphics[height=0.2\linewidth]
			{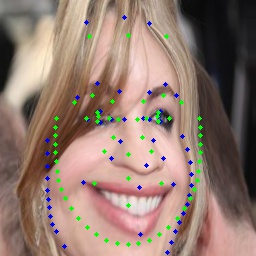}&
            \includegraphics[height=0.2\linewidth]
			{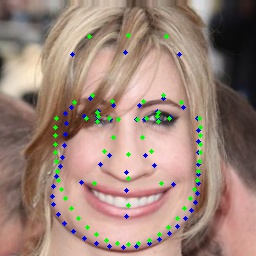}&
            \includegraphics[height=0.2\linewidth]
			{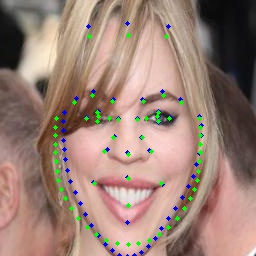}\\
            \includegraphics[height=0.2\linewidth]
			{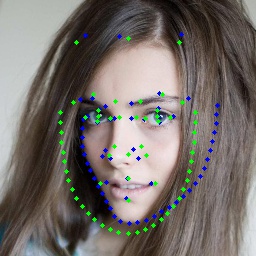}&
            \includegraphics[height=0.2\linewidth]
			{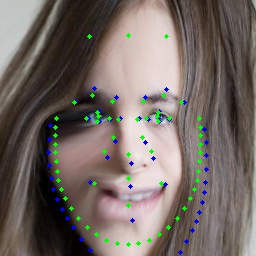}&
            \includegraphics[height=0.2\linewidth]
			{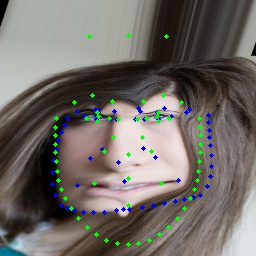}&
            \includegraphics[height=0.2\linewidth]
			{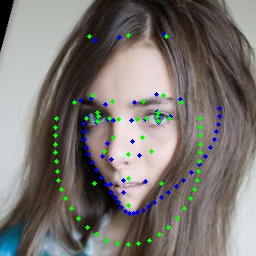}\\
            Input& $\mathcal{L}_{\text{adv}}$&$\mathcal{L}_{\text{adv}}+\lambda_{\text{cyc}} \mathcal{L}_{\text{cyc}}$&$\mathcal{L}_{geo}$\\
		\end{tabular}}

    \caption{Comparing our \emph{CariGeoGAN} with different losses. The green points represent the landmarks of the mean face, while the blue ones represent the landmarks of the input or exaggerated face. Input images: CelebA dataset.}
	\label{fig:geo_loss_compare}

\end{figure}

\begin{figure}
	\footnotesize
	\setlength{\tabcolsep}{0.003\linewidth}
    \scalebox{1.1}{
		\begin{tabular}{cccc}
            \includegraphics[height=0.2\linewidth]
			{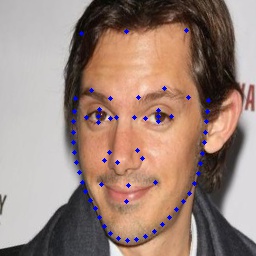}& 
            \includegraphics[height=0.2\linewidth]
			{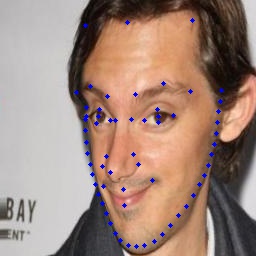}&
            \includegraphics[height=0.2\linewidth]
			{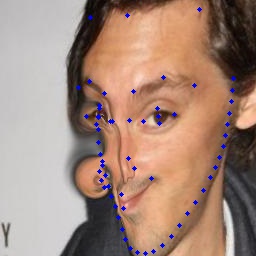}&
            \includegraphics[height=0.2\linewidth]
			{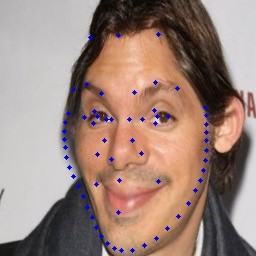}\\ 
            \includegraphics[height=0.2\linewidth]
			{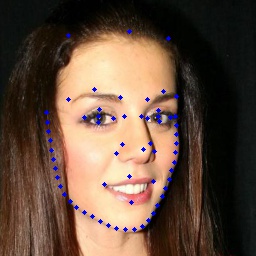}& 
            \includegraphics[height=0.2\linewidth]
			{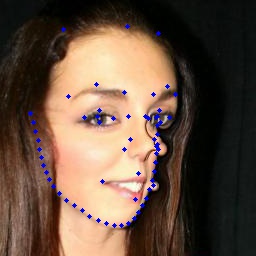}&
            \includegraphics[height=0.2\linewidth]
			{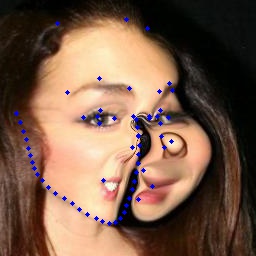}&
            \includegraphics[height=0.2\linewidth]
			{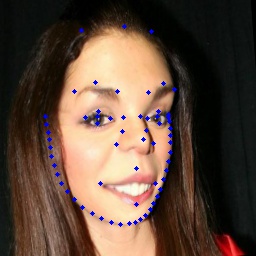}\\        
             \includegraphics[height=0.2\linewidth]
			{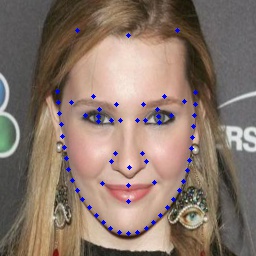}& 
            \includegraphics[height=0.2\linewidth]
			{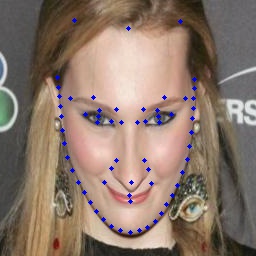}&
            \includegraphics[height=0.2\linewidth]
			{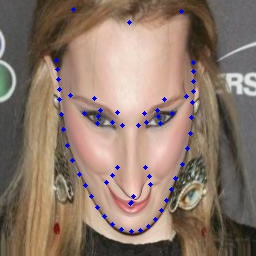}&
            \includegraphics[height=0.2\linewidth]
			{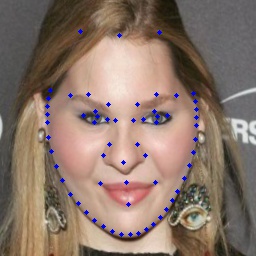}\\        
            
            Input&  factor$ = 2$ & factor = $ 3$ & \emph{CariGeoGAN}
		\end{tabular}}
	
    \caption{Comparing our \emph{CariGeoGAN} to simple exaggeration of the PCA coefficients from the mean by a factor of $2$ or $3$. Input images: CelebA dataset.}
    \vspace{-0.1in}
	\label{fig:from_mean}
\end{figure}

Our objective function for optimizing \emph{CariGeoGAN} is:
\begin{align}
\mathcal{L}_{geo} = &\mathcal{L}_{\text{adv}}^{L_X}+ \mathcal{L}_{\text{adv}}^{L_Y} + \lambda_{\text{cyc}} \mathcal{L}_{\text{cyc}} + \lambda_{\text{cha}} (\mathcal{L}_{\text{cha}}^{L_X} + \mathcal{L}_{\text{cha}}^{L_Y}),
\end{align}
where $\lambda_{cyc}$ and $\lambda_{cha}$ balance the multiple objectives. 

\fref{fig:geo_loss_compare} shows the roles of each loss, which is added to the objective function one by one. With adversarial only, the model will collapse and all face shapes in $L_X$ map to a very similar caricature shape. With adding Cycle-consistency loss, the output varies with the input but the exaggeration direction is arbitrary. By adding characteristic loss, the exaggeration becomes meaningful. It captures the most distinct face features compared with the mean face, and then exaggerates the distinct facial features. Please note although the characteristic loss encourages the direction but itself is not enough to determine the exaggeration, since it cannot constrain the exaggeration amplitude and relationship between different facial features. For example, if we simply amplify differences from the mean by a factor of $2$ or $3$, it minimizes $\mathcal{L}_{\text{cha}}$ but leads to unsatisfactory results as shown in \fref{fig:from_mean}. In summary, our geometric exaggeration is learned from data by \emph{CariGeoGAN} to balance all the four types of losses. That is also the major difference compared to hand-crafted rules used in previous EDFM methods.

\begin{figure*}
	\footnotesize
	\setlength{\tabcolsep}{0.003\linewidth}
    \scalebox{1.1}{
		\begin{tabular}{cccccccc}
        	
            \includegraphics[height=0.105\textwidth]
			{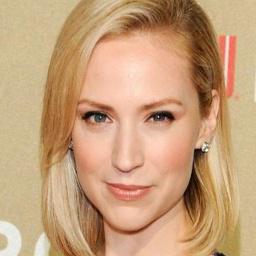}&
            \includegraphics[height=0.105\textwidth]
			{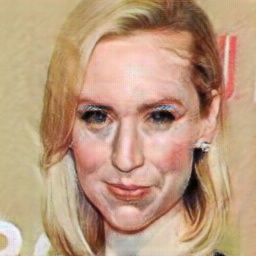}&
			\includegraphics[height=0.105\textwidth]
			{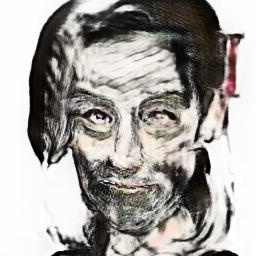}&
            \includegraphics[height=0.105\textwidth]
			{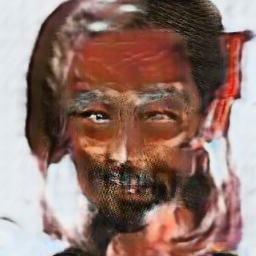}&
            \includegraphics[height=0.105\textwidth]
			{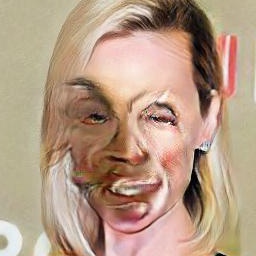}&
			\includegraphics[height=0.105\textwidth]
			{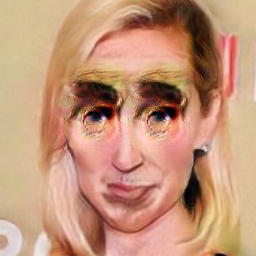}&
			\includegraphics[height=0.105\textwidth]
			{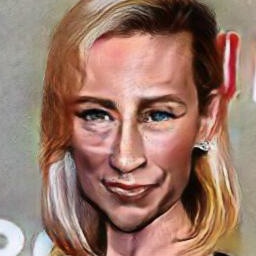}&
			\includegraphics[height=0.105\textwidth]
			{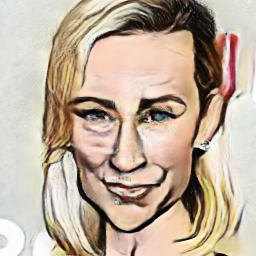}
            \\
			\includegraphics[height=0.105\textwidth]
			{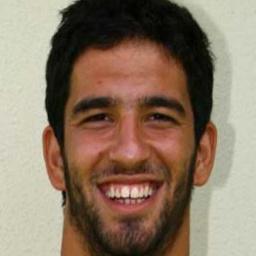}&
			\includegraphics[height=0.105\textwidth]
			{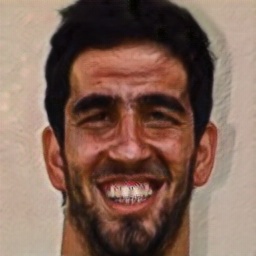}&
			\includegraphics[height=0.105\textwidth]
			{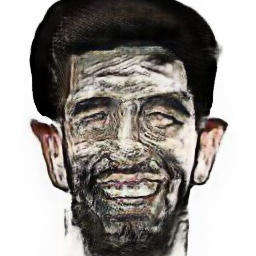}&
			\includegraphics[height=0.105\textwidth]
			{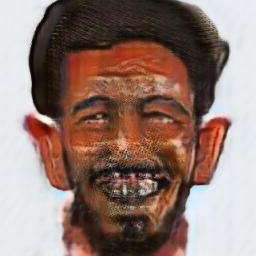}&
			\includegraphics[height=0.105\textwidth]
			{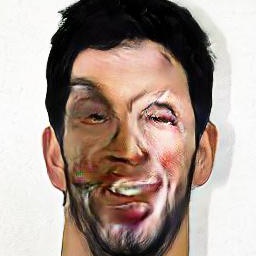}&
			\includegraphics[height=0.105\textwidth]
			{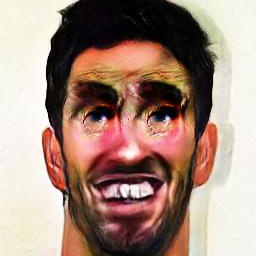}&
			\includegraphics[height=0.105\textwidth]
			{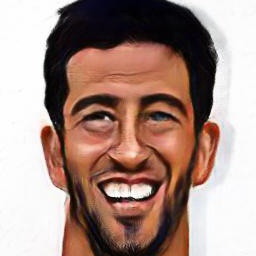}&
			\includegraphics[height=0.105\textwidth]
			{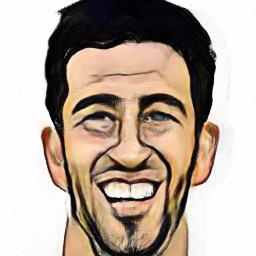}
			\\
			\includegraphics[height=0.105\textwidth]
			{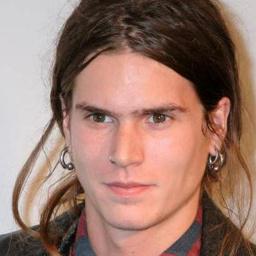}&
			\includegraphics[height=0.105\textwidth]
			{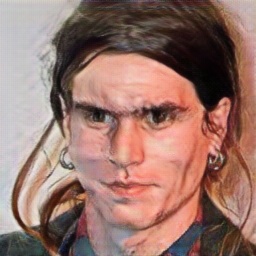}&
			\includegraphics[height=0.1\textwidth]
			{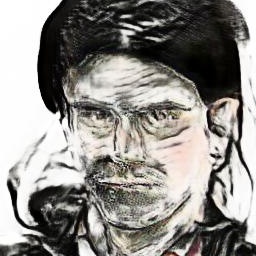}&
			\includegraphics[height=0.105\textwidth]
			{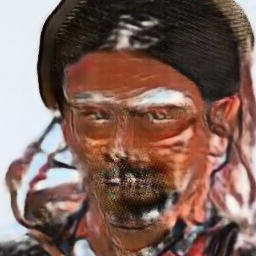}&
			\includegraphics[height=0.105\textwidth]
			{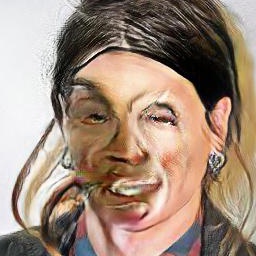}&
			\includegraphics[height=0.105\textwidth]
			{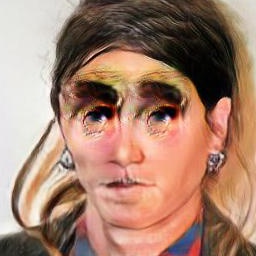}&
			\includegraphics[height=0.105\textwidth]
			{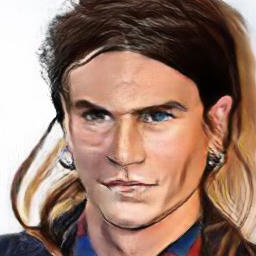}&
			\includegraphics[height=0.105\textwidth]
			{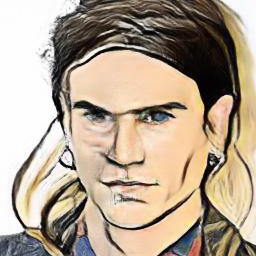}\\
            Input&CycleGAN&MUNIT &MUNIT &Ours w/o $\mathcal{L}_{\text{per}}$&Ours w/o $\mathcal{L}_{\text{per}}$ &Ours  & Ours
        	\\
		\end{tabular}}

    \caption{Comparing our \emph{CariStyGAN} with CycleGAN\citep{zhu2017toward} and MUNIT\citep{huang2018munit}. All networks are trained with the same datasets to learn appearance style mapping $X \Rightarrow Y'$. CycleGAN generates a single result (2nd column). MUNIT is capable to generate diverse results but fails to preserve face structure (3rd and 4th second columns). Our \emph{CariStyGAN} generates better diverse results by combining both CycleGAN and MUNIT (5th and 6th columns), and preserves identity by introducing a new perceptual loss (7th and 8th columns). Input images: CelebA dataset.}
	\label{fig:appear_compare_exm1}
\end{figure*}

\paragraph{\textbf{Training details}}
We use the same training strategy to CycleGAN~\cite{CycleGAN2017}. For all the experiments, we set $\lambda_{cyc}=10$ and $\lambda_{cha}=1$ empirically and use the Adam solver~\cite{kingma2014adam} with a batch size of $1$. All networks are trained from scratch with an initial learning rate of $0.0002$.

\subsection{Appearance Stylization}

In this section, we present \emph{CariStyGAN}, which learns to apply appearance styles of caricatures to portrait photos without changes in geometry.

\paragraph{\textbf{Training data}}
In order to learn a pure appearance stylization without any geometric deformation, we need to synthesize an intermediate domain $Y'$, which has the same geometry distribution to $X$ and has the same appearance distribution as $Y$. We synthesize each image $\{y'_i\}_{i={1,...,M}}, y'_i \in Y'$ by warping every caricature image $y_i \in Y$ with the landmarks translated by our \emph{CariGeoGAN}: $G_{L_X}(y_i)$. Our \emph{CariStyGAN} learns the translation from $X$ to $Y'$ instead.

\begin{figure*}
\centering
\includegraphics[width=\textwidth]{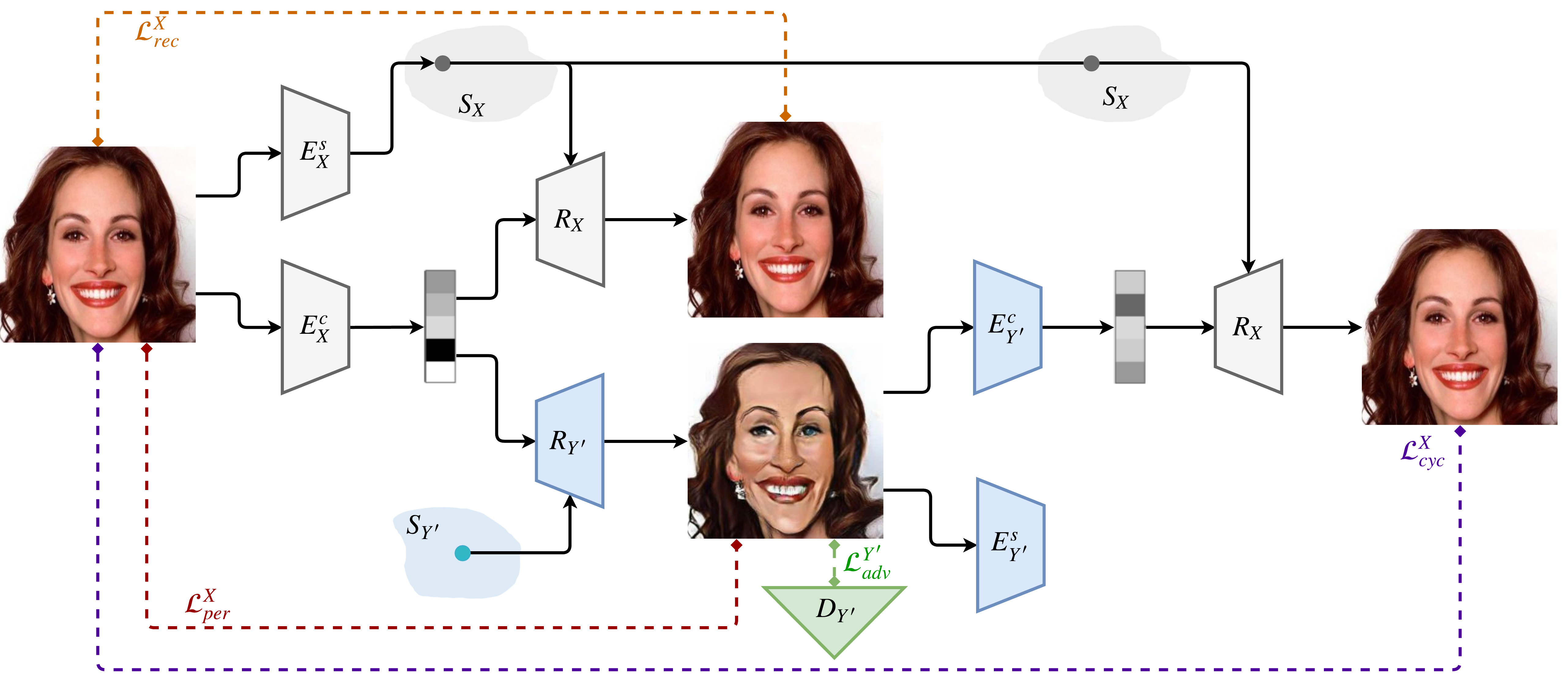}
\caption{Architecture of our \emph{CariStyGan}. For simplicity, here we only show the network architecture for the translation $X \rightarrow Y'$. And the network architecture for the reverse translation $Y' \rightarrow X$ is symmetric. Input image: CelebA dataset.}
\vspace{-0.1in}
\label{fig:appearance}
\end{figure*}

\paragraph{\textbf{CariStyGAN}}
Since the mapping from $X$ to $Y'$ is a typical image-to-image translation task without geometric deformation, some general image-to-image translation network, \eg, CycleGAN~\cite{CycleGAN2017}, MNUIT~\cite{huang2018munit}, can be applied. As shown in the 2nd, 3rd and 4th columns of \fref{fig:appear_compare_exm1}, the result obtained by CycleGAN is acceptable in preserving structure but lacks diversity; while MUNIT generates multiple results with various styles but these results fail to preserve face structure. We find the reason is that the feature-level cycle-consistency used in MUNIT is less constrained than the image-level cycle-consistency used in CycleGAN. This is verified by replacing the feature-level cycle-consistency in MUNIT with the image-level one, results of which are shown in 5th and 6th columns of \fref{fig:appear_compare_exm1}. 

Our \emph{CariStyGan} combines merits of the two networks, \ie, allowing diversity and preserving structure. We inherit the image-level cycle-consistency constraint from CycleGAN to keep the face structure, while we are inspired from MNUIT to explicitly disentangle image representation into a content code that is domain-invariant, and a style code that captures domain-specific properties. By recombining a content code with various style codes sampled from the style space of the target domain, we may get multiple translated results.

Different from a traditional auto-encoder structure, we design an auto-encoder consisting of two encoders and one decoder for images from each domain $I (I = X, Y')$. The content encoder $E_I^c$ and the style encoder $E_I^s$, factorize the input image $z_I \in I$ into a content code $c_I$ and a style code $s_I$ respectively, \ie, $(c_I,s_I)=(E_I^c(z_I),E_I^{s}(z_I))$. The decoder $R_I$ reconstructs the input image from its content and style code, $z_I=R_I(c_I,s_I)$. The domain of style code $S_I$ is assumed to be Gaussian distribution $\mathcal{N}(0,1)$. 

\fref{fig:appearance} shows our network architecture in the forward cycle. Image-to-image translation is performed by swapping the encoder-decoder pairs of the two domains. For example, given a portrait photo $x\in X$, we first extract its content code $c_x = E_X^c(x)$ and randomly sample a style code $s_{y'}$ from $S_{Y'}$. Then, we use the decoder $R_{Y'}$ instead of its original decoder $R_X$ to produce the output image $y'$, in caricature domain $Y'$, denoted as $y'=R_{Y'}(c_x,s_{y'})$. $y'$ is also constrained by the discriminator $D_{Y'}$.

By contrast to MUNIT~\cite{huang2018munit}, where the cycle-consistency is enforced in the two code domains, we enforce cycle-consistency in the image domain. It means that the recovered image $\widehat{x} = R_{X}(E_{Y'}^c(y'))$ should be close to the original input $x$.

With this architecture, the forward mapping $\Phi_{app}: X \rightarrow Y'$ is achieved by $E_{X}+R_{Y'}$, while the back mapping $\Phi^{-1}_{app}: Y' \rightarrow X$ is achieved by $E_{Y'}+R_{X}$. By sampling different style codes, the mappings become multi-modal.

\paragraph{\textbf{Loss}}
The \emph{CariStyGAN} comprises four types of loss, which are shown in \fref{fig:appearance}.

The first is adversarial loss $\mathcal{L}_{\text{adv}}^{Y'}(E_X,R_{Y'},D_{Y'})$, which makes the translated result $R_{Y'}(E_X^c(x),s_{y'})$ identical to the real sample in $Y'$, where $D_{Y'}$ is a discriminator to distinguish the generated samples from the real ones in $Y'$. Another adversarial loss $\mathcal{L}_{\text{adv}}^{X}(E_{Y'},R_{X},D_{X})$ is similarly defined for the reverse mapping $Y'\rightarrow X$, where $D_{X}$ is discriminator for $X$.

The second is reconstruction loss which penalizes the $L1$ differences between the input image and the result, reconstructed from its style code and content code, \ie,
\begin{equation}
\mathcal{L}_{\text{rec}}^{I}(E^c_{I},E^s_{I},R_{I}) =  \mathbb{E}_{z \sim I}[||R_I(E_I^c(z),E_I^s(z))-z||_1]
\label{eq:rec}
\end{equation}

The third is cycle-consistency loss, which enforces the image to get back after forward and backward mappings. Specifically, given $x \in X$, we get the result $R_{Y'}(E_X^c(x),s_{y'}), s_{y'} \in S_{Y'},$ which translates from $X$ to $Y'$. The result is then fed into the encoder $E_{Y'}^c$ to get its content code. After combining the content code with a random code $s_x$ sampled form $S_X$, we use the decoder $R_X$ to get the final result. It should be the same to the original input $x$: 
\begin{align}
&\mathcal{L}_{\text{cyc}}^X(E^c_X,R_{Y'},E^c_{Y'},R_{X}) = \nonumber\\
&\mathbb{E}_{x \sim X', s_x \sim S_X,s_{y'} \sim S_{Y'}} [||R_X(E^c_{Y'}(R_{Y'}(E_X^c(x),s_{y'})),s_x)-x||_{1} ] ,
\label{eq:per}
\end{align}
The cycle loss for the reverse mapping $Y'\rightarrow X$ is defined symmetrically and denoted as $\mathcal{L}_{\text{cyc}}^{Y'}(E^c_{Y'},R_{X},E^c_{X},R_{Y'})$.

The aforementioned three losses are inherited form MUNIT and cycleGAN. With the three only, the decoupling of style code and content code is implicitly learned and vaguely known. In the photo-caricature task, we find that style code is not completely decoupled with the content code, which may cause the failure of preserving the identity after translation, as shown in the 5$\&6-th$ rows of \fref{fig:appear_compare_exm1}. To address this issue, we add a new perceptual loss \cite{johnson2016perceptual}, which can explicitly constrain the translated result to have the same content information to the input:
\begin{align}
\mathcal{L}_{\text{per}}^X(E^c_{X},R_{Y'}) =&  \mathbb{E}_{x \sim X,s_{y'} \sim Y'}[||\text{VGG19}_{5\_3}(R_{Y'}(E_X^c(x),s_{y'})) \nonumber\\
-&\text{VGG19}_{5\_3}(x)||_2],
\label{eq:per}
\end{align}
where $\text{VGG19}_{5\_3}$ denotes to the $relu5\_3$ feature map in VGG19 \citep{simonyan2014very}, pre-trained on image recognition task. $\mathcal{L}_{\text{per}}^{Y'}(E^c_{Y'},R_{X})$ is defined symmetrically. \vspace{1.0em}

In summary, we jointly train the encoders, decoders and discriminators in our \emph{CariStyGAN} by optimizing the final loss function:
\begin{align}
\mathcal{L}_{app} = &\mathcal{L}_{\text{adv}}^{X}+ \mathcal{L}_{\text{adv}}^{Y'} + \lambda_{\text{rec}} (\mathcal{L}_{\text{rec}}^X + \mathcal{L}_{\text{rec}}^{Y'}) + \lambda_{\text{cyc}} (\mathcal{L}^{X}_{\text{cyc}}+ \mathcal{L}_{\text{cyc}}^{Y'}) \nonumber \\
+ & \lambda_{\text{per}} (\mathcal{L}_{\text{per}}^{X} + \mathcal{L}_{\text{per}}^{Y'}).
\end{align}
$\lambda_{rec}$, $\lambda_{cyc}$ and $\lambda_{per}$ balance the multiple objectives.

\paragraph{\textbf{Training details}}
We use the same structure as MUNIT in our encoders, decoders, and discriminators, and follow its training strategy. For all the experiments, we set $\lambda_{rec}=1$, $\lambda_{per}=0.2$ and $\lambda_{cyc}=1$ empirically and use the Adam solver~\cite{kingma2014adam} with a batch size of $1$. All networks are trained from scratch with an initial learning rate of $0.0001$.

\section{Discussion}
In this section, we analyze two key components in our \emph{CariGANs}, \ie, PCA representation in \emph{CariGeoGAN} and intermediate domain in \emph{CariStyGAN}.
\begin{figure}
\centering
\includegraphics[width=0.5\textwidth]{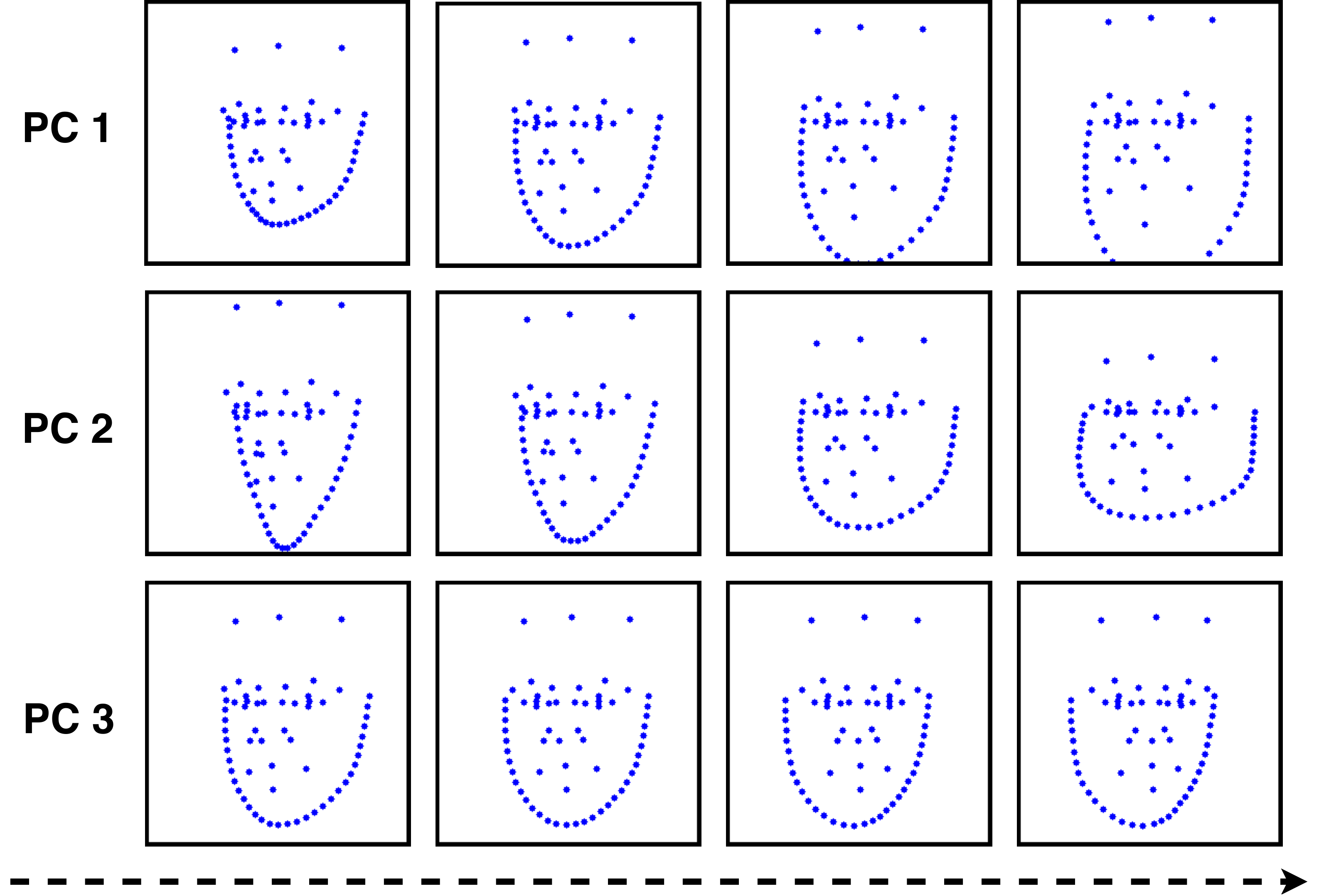}
\vspace{-0.1in}
\caption{Visualization of top three principal components of landmarks}
\label{fig:PCA}
\end{figure}

\begin{figure}
	\footnotesize
	\setlength{\tabcolsep}{0.003\linewidth}
    \scalebox{1.1}{
		\begin{tabular}{ccc}
           \includegraphics[height=0.29\linewidth]
			{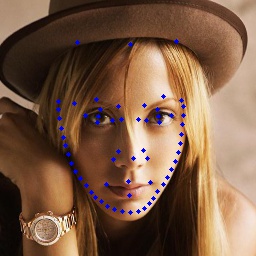}&
           \includegraphics[height=0.29\linewidth]
			{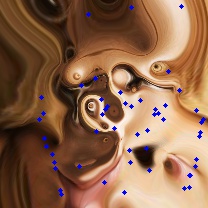}&
            \includegraphics[height=0.29\linewidth]
			{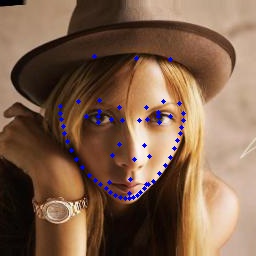}
        	\\
            \includegraphics[height=0.29\linewidth]
			{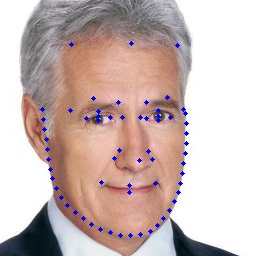}&
            \includegraphics[height=0.29\linewidth]
			{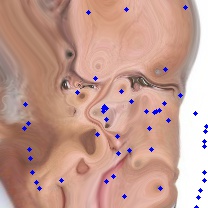}&
            \includegraphics[height=0.29\linewidth]
			{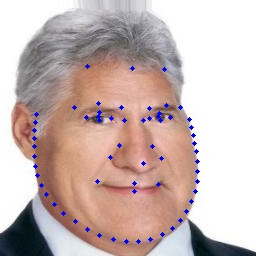}\\
            Input&Result with 2D coords & Result with PCA
		\end{tabular}}

    \caption{Comparison between using PCA representation and using 2D coordinate representation in \emph{CariGeoGAN}. Input images: CelebA dataset.}
    \vspace{-0.1in}
	\label{fig:point_vs_pca}
\end{figure}

\subsection{Why PCA representation is essential to \emph{CariGeoGAN}?}
\label{sec:pca}
Although the dimensionality of landmarks with 2D coordinates ($63 \times 2$) is low compared to the image representation, in our \emph{CariGeoGAN} we still use the PCA to reduce dimensions of the landmark representation. That is because geometry translation is sometimes harder than image translation. First, landmarks are fed into fully-connected layers instead of convolutional layers, so they lose the locally spatial constraint during learning. Second, the result is more sensitive to small errors in landmarks than in image pixels, since these errors may cause serious geometric artifacts, like foldover or zigzag contours. If we use the raw 2D coordinates of landmarks to train \emph{CariGeoGAN}, the face structure is hardly preserved as shown in \fref{fig:point_vs_pca}. On the contrary, the PCA helps constrain the face structure in the output. It constructs an embedding space of face shapes, where each principle component represents a direction of variants, like pose, shape, size, as shown by the visualization of top three principle components in \fref{fig:PCA}. Any sample in the embedding space will maintain the basic face structure. We compared two kinds of representations used in \emph{CariGeoGAN} (PCA \emph{vs.} 2D coordinate), and show visual results in \fref{fig:point_vs_pca}.

\subsection{Why is intermediate domain crucial to \emph{CariStyGAN}?}
\begin{figure}
	\footnotesize
	\setlength{\tabcolsep}{0.003\linewidth}
    \scalebox{1.1}{
		\begin{tabular}{cccc}
            \includegraphics[height=0.29\linewidth]
			{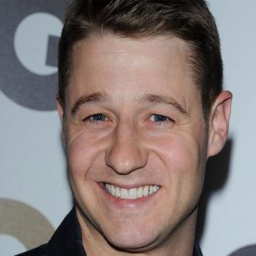}&        
           \includegraphics[height=0.29\linewidth]
			{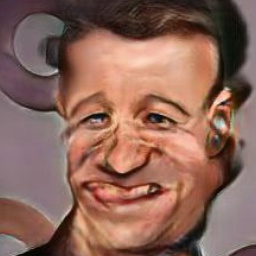}&
            \includegraphics[height=0.29\linewidth]
			{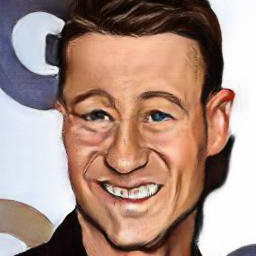}\\
            \includegraphics[height=0.29\linewidth]
			{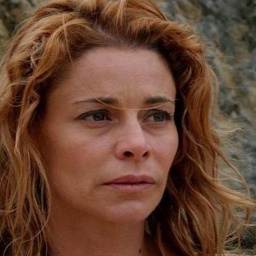}&       
           \includegraphics[height=0.29\linewidth]
			{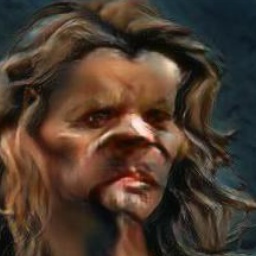}&
            \includegraphics[height=0.29\linewidth]
			{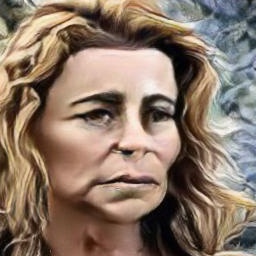}\\    
                   \includegraphics[height=0.29\linewidth]
			{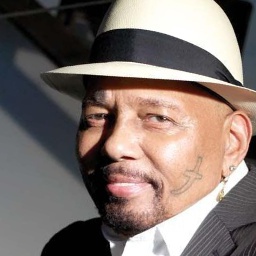}&       
           \includegraphics[height=0.29\linewidth]
			{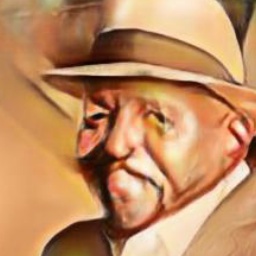}&
            \includegraphics[height=0.29\linewidth]
			{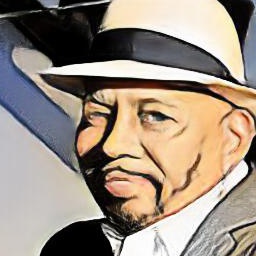}\\  
            Input& $X\rightarrow Y$ &  $X\rightarrow Y'$ 
		\end{tabular}}
	\vspace{-0.1in}
    \caption{The importance of intermediate domain to train \emph{CariStyGAN}. Input images: CelebA dataset.}
    \vspace{-0.1in}
	\label{fig:interme}
\end{figure}

The construction of the intermediate domain $Y'$ is an important factor for the success of our \emph{CariStyGAN}, since it bridges the geometric differences between photo domain $X$ and caricature domain $Y$, and thus allows the GAN focusing on appearance translation only. To understand the role of $Y'$ well, we train \emph{CariStyGAN} to learn the mapping from $X$ to $Y$ directly. In this setting, some textures may mess up the face structure, as shown in \fref{fig:interme}. One possible reason is that the network attempts to learn two mixed mappings (geometry and appearance) together. The task is so difficult to be learned.

\subsection{How many styles have been learned by \emph{CariStyGAN}?}

\begin{figure*}
\centering
\includegraphics[width=0.95\textwidth]{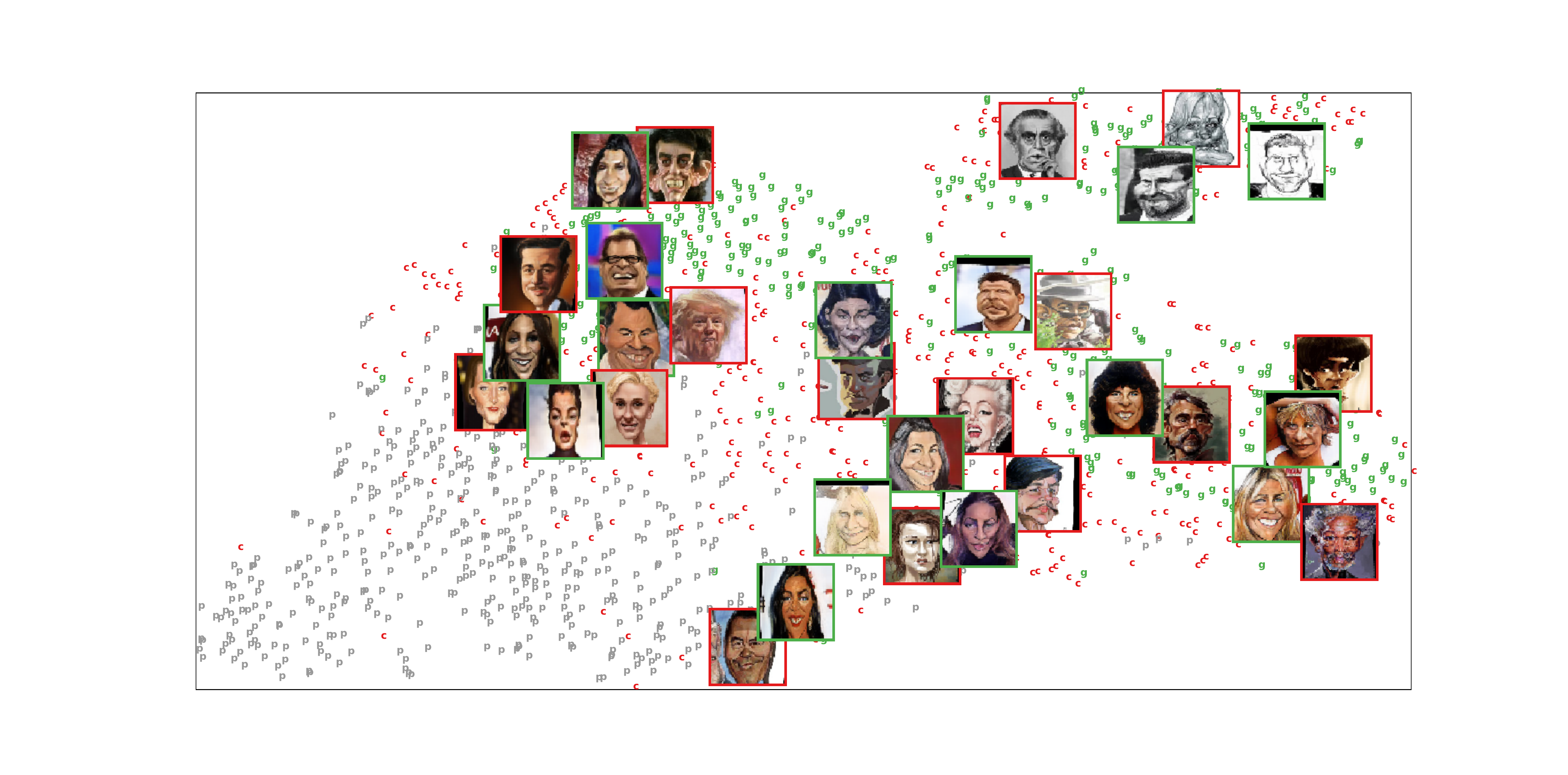}
\vspace*{-0.1in}
\caption{T-SNE visualization of the style embedding. Gray points represent photos, red points represent hand-drawn caricatures, and green points represent generated results. The corresponding image of some point is shown with the same color border.}

\label{fig:tsne}

\end{figure*}

To answer this question, we randomly select $500$ samples from the testing photo dataset and $500$ samples from the of training hand-drawn caricatures dataset. For each photo sample we generate a caricature with our \emph{CariGANs} and a random style code. Then we follow \cite{gatys2015neural} to represent the appearance style of each sample with the Gram Matrix of its VGG19 feature maps. The embedding of appearance styles on 2D is visualized via the T-SNE method. It is clearly shown in \fref{fig:tsne}, there is little interaction between photos and hand-drawn caricatures, however, through translation our generated results almost share the same embedding space as caricatures. That means most styles in the training caricature dataset have been learned by our \emph{CariStyGAN}.

\section{Comparison and Result}
In this section, we first show the performance of our system, and demonstrate the result controllability in two aspects. Then we qualitatively compare our results to previous techniques, including both traditional graphics-based methods and recent deep-learning based methods. Finally, we provide the perceptual study results.

\subsection{Performance}
Our core algorithm is developed in PyTorch ~\cite{paszke2017automatic}. All of our experiments are conducted on a PC with an Intel E5 2.6GHz CPU and an NVIDIA K40 GPU. The total runtime for a $256 \times 256$ image is approximately 0.14 sec., including 0.10 sec. for appearance stylization, 0.02 sec. for geometric exaggeration and 0.02 sec. for image warping.

\subsection{Results with control}

\begin{figure}
	\footnotesize
	\setlength{\tabcolsep}{0.003\linewidth}
    \scalebox{1.1}{
		\begin{tabular}{cccc}
			\includegraphics[height=0.21\linewidth]
			{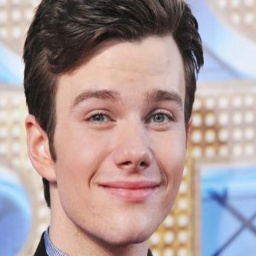}&
            \includegraphics[height=0.21\linewidth]
			{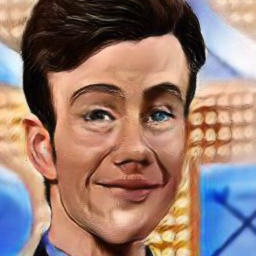}&
            \includegraphics[height=0.21\linewidth]
			{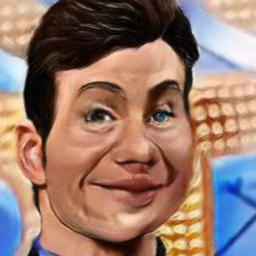}&
            \includegraphics[height=0.21\linewidth]
			{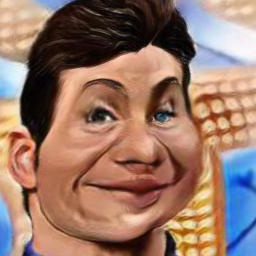}
            \\
            \includegraphics[height=0.21\linewidth]
			{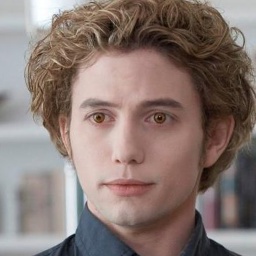}&
            \includegraphics[height=0.21\linewidth]
			{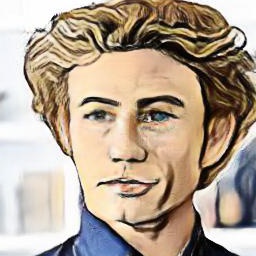}&
            \includegraphics[height=0.21\linewidth]
			{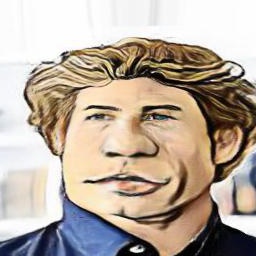}&
            \includegraphics[height=0.21\linewidth]
			{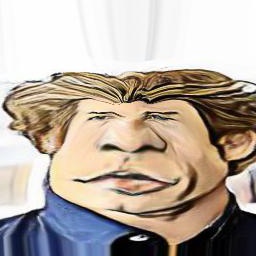}
            \\
            \includegraphics[height=0.21\linewidth]
			{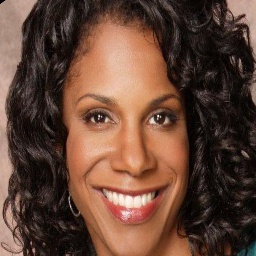}&
            \includegraphics[height=0.21\linewidth]
			{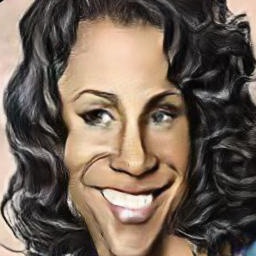}&
            \includegraphics[height=0.21\linewidth]
			{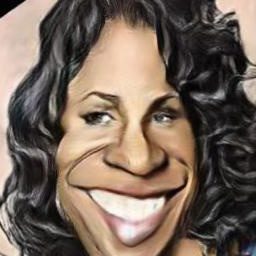}&
            \includegraphics[height=0.21\linewidth]
			{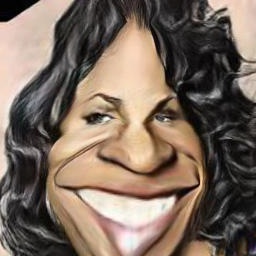}
            \\
            Input&$\alpha=0.0$&$\alpha=1.0$&$\alpha=2.0$
		\end{tabular}}
	\vspace{-0.1in}
    \caption{Results with geometric control. Inputs are from CelebA dataset excluding the 10K images used in training.}
	\label{fig:geo_control}
\end{figure}

Our \emph{CariGANs} support two aspects of control. First, our system allows users to tweak the geometric exaggeration extent with a parameter $\alpha \in [0.0,2.0]$. Let $l_x$ to be the original landmarks of the input photo $x$, and $l_y$ to be the exaggerated landmarks predicted by \emph{CariGeoGAN}. Results with different exaggeration degrees can be obtained by interpolation and extrapolation between them: $l_x+\alpha(l_y-l_x)$. \Fref{fig:geo_control} shows such examples.

\begin{figure*}
	\footnotesize
	\setlength{\tabcolsep}{0.002\linewidth}
    \scalebox{1.1}{
		\begin{tabular}{ccccccc}
        			&
			&
			&
			&
			&
			\includegraphics[height=0.125\textwidth]
			{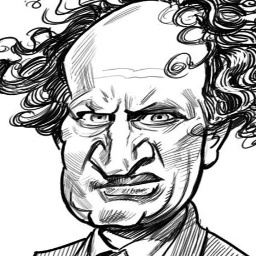}&
			\includegraphics[height=0.125\textwidth]
			{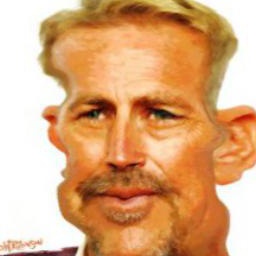}\\
			\includegraphics[height=0.125\linewidth]
			{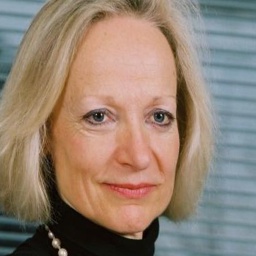}&
			\includegraphics[height=0.125\linewidth]
			{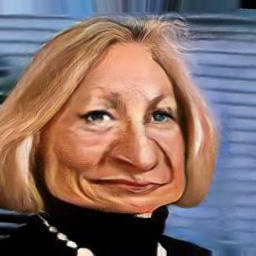}&
			\includegraphics[height=0.125\textwidth]
			{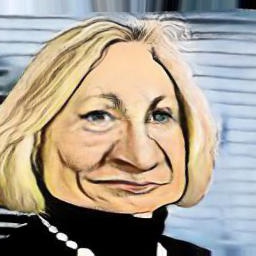}&
			\includegraphics[height=0.125\textwidth]
			{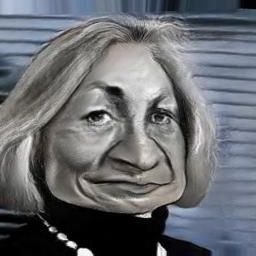}&
			\includegraphics[height=0.125\linewidth]
			{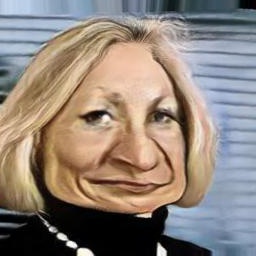}&
			\includegraphics[height=0.125\textwidth]
			{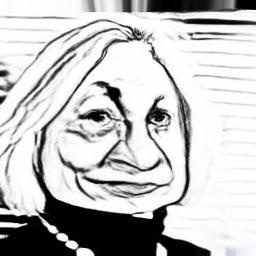}&
			\includegraphics[height=0.125\textwidth]
			{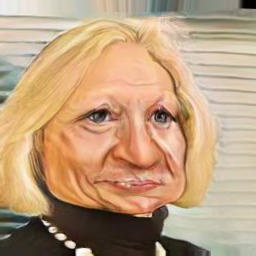}\\
            \includegraphics[height=0.125\linewidth]
			{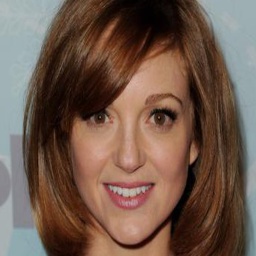}&
			\includegraphics[height=0.125\linewidth]
			{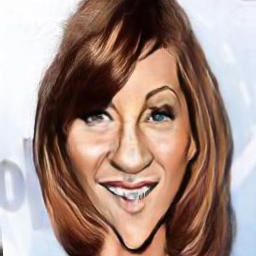}&
			\includegraphics[height=0.125\textwidth]
			{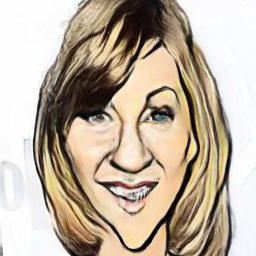}&
			\includegraphics[height=0.125\textwidth]
			{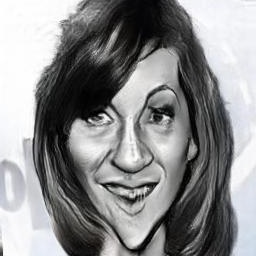}&
			\includegraphics[height=0.125\linewidth]
			{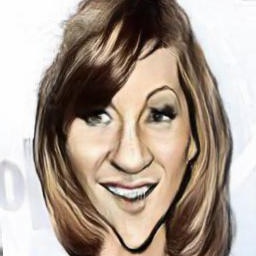}&
			\includegraphics[height=0.125\textwidth]
			{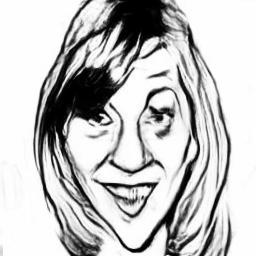}&
			\includegraphics[height=0.125\textwidth]
			{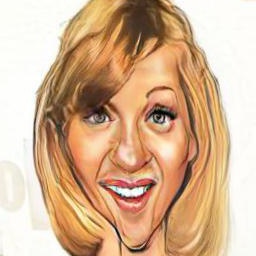}\\
                    \includegraphics[height=0.125\linewidth]
			{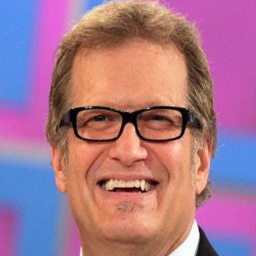}&
			\includegraphics[height=0.125\linewidth]
			{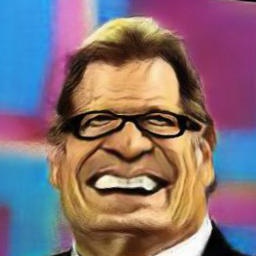}&
			\includegraphics[height=0.125\textwidth]
			{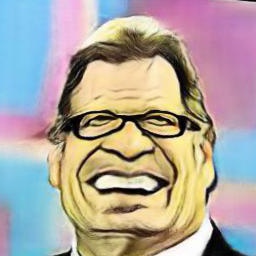}&
			\includegraphics[height=0.125\textwidth]
			{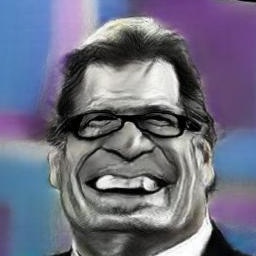}&
			\includegraphics[height=0.125\linewidth]
			{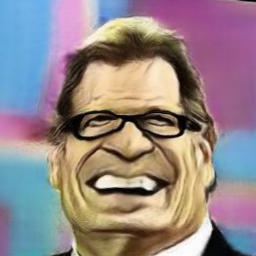}&
			\includegraphics[height=0.125\textwidth]
			{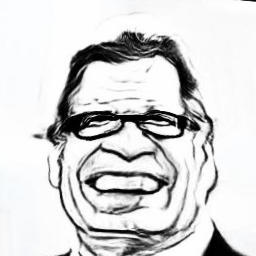}&
			\includegraphics[height=0.125\textwidth]
			{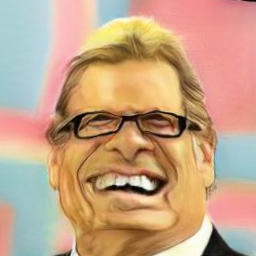}\\
                   \includegraphics[height=0.125\linewidth]
			{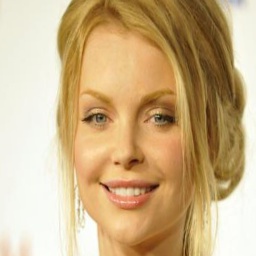}&
			\includegraphics[height=0.125\linewidth]
			{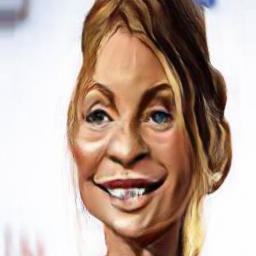}&
			\includegraphics[height=0.125\textwidth]
			{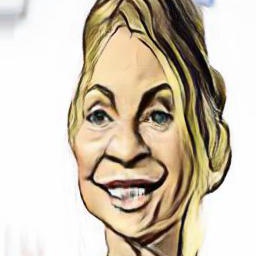}&
			\includegraphics[height=0.125\textwidth]
			{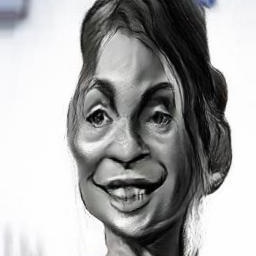}&
			\includegraphics[height=0.125\linewidth]
			{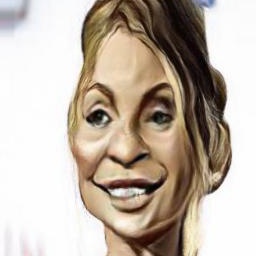}&
			\includegraphics[height=0.125\textwidth]
			{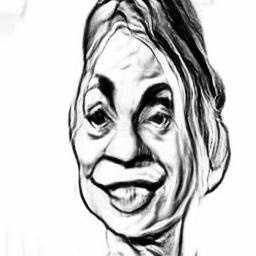}&
			\includegraphics[height=0.125\textwidth]
			{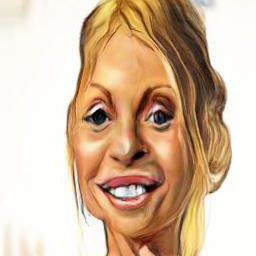}\\
            
             \includegraphics[height=0.125\linewidth]
			{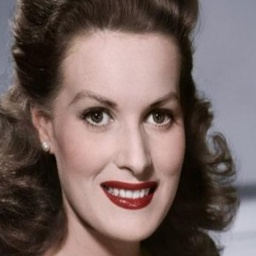}&
			\includegraphics[height=0.125\linewidth]
			{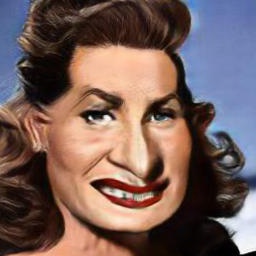}&
			\includegraphics[height=0.125\textwidth]
			{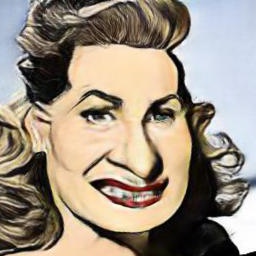}&
			\includegraphics[height=0.125\textwidth]
			{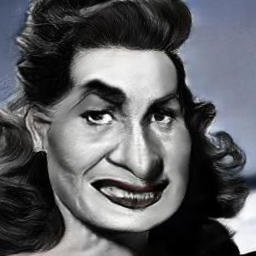}&
			\includegraphics[height=0.125\linewidth]
			{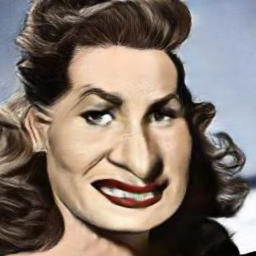}&
			\includegraphics[height=0.125\textwidth]
			{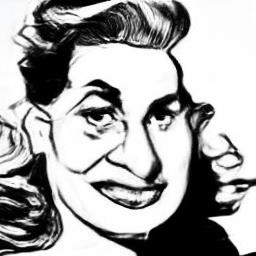}&
			\includegraphics[height=0.125\textwidth]
			{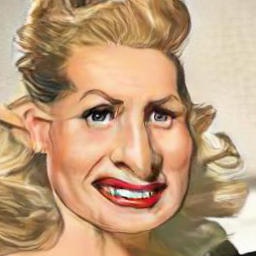}\\
              \includegraphics[height=0.125\linewidth]
			{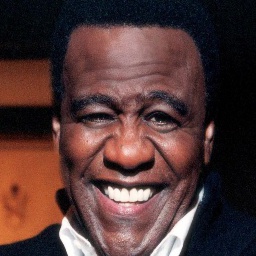}&
			\includegraphics[height=0.125\linewidth]
			{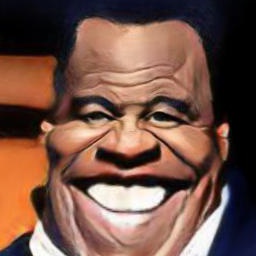}&
			\includegraphics[height=0.125\textwidth]
			{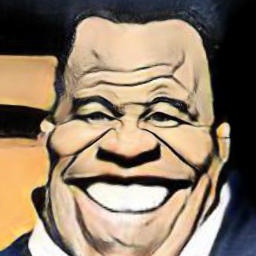}&
			\includegraphics[height=0.125\textwidth]
			{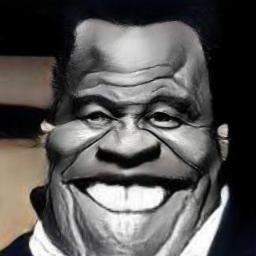}&
			\includegraphics[height=0.125\linewidth]
			{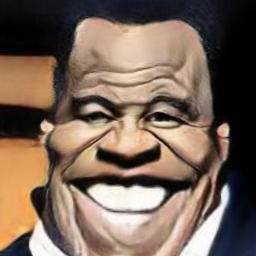}&
			\includegraphics[height=0.125\textwidth]
			{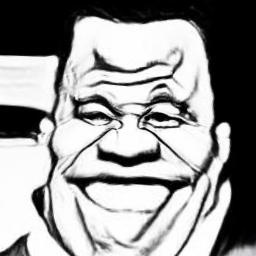}&
			\includegraphics[height=0.125\textwidth]
			{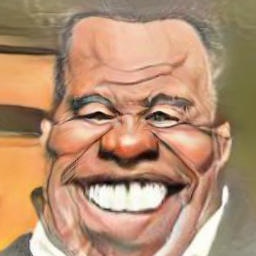}\\
          \includegraphics[height=0.125\linewidth]
			{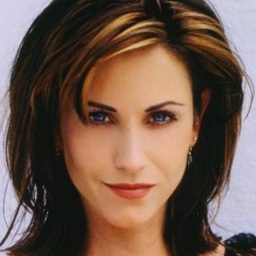}&
			\includegraphics[height=0.125\linewidth]
			{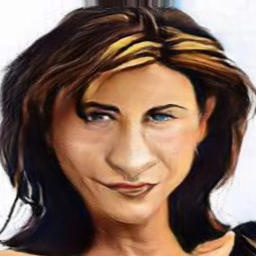}&
			\includegraphics[height=0.125\textwidth]
			{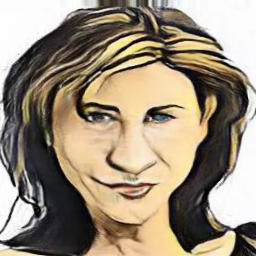}&
			\includegraphics[height=0.125\textwidth]
			{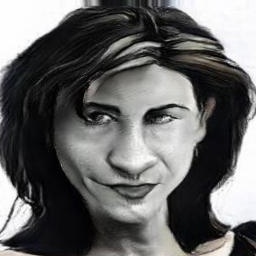}&
			\includegraphics[height=0.125\linewidth]
			{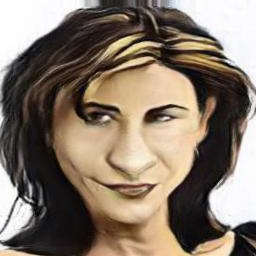}&
			\includegraphics[height=0.125\textwidth]
			{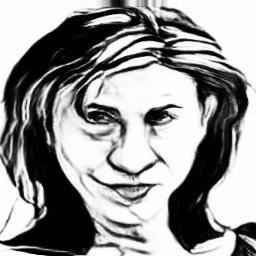}&
			\includegraphics[height=0.125\textwidth]
			{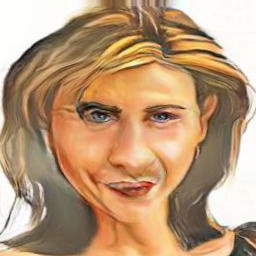}\\
			Input& Result with code1 & Result with code2 & Result with code3 & Result with code4 & result with ref1 & Result with ref2 
		
		\end{tabular}}
	
    \caption{Our system allows user control on appearance style. Results are generated with a random style code (first four) or a given reference (last two). Top row shows the two reference images. From left to right: \textcopyright Tom Richmond/tomrichmond.com, \textcopyright wooden-horse/deviantart. Inputs are from CelebA dataset excluding the 10K images used in training.}
	\label{fig:sty_control}
\end{figure*}

Except for geometry, our system allows user control on appearance style as well. On one hand, our \emph{CariStyGAN} is a multi-modal image translation network, which can convert a given photo into different caricatures, which are obtained by combining photos with different style codes, sampled from a Gaussian distribution. On the other hand, in the \emph{CariStyGAN}, a reference caricature can be encoded into style code using $E^s_{Y'}$. After combining with the code, we can get the result with a similar style as the reference. So the user can control the appearance style of the output by either tuning the value of style code or giving a reference. In \Fref{fig:sty_control}, we show diverse results with 4 random style codes and 2 style codes from references.

\begin{figure*}
	\footnotesize
	\setlength{\tabcolsep}{0.002\linewidth}
    \scalebox{1.1}{
		\begin{tabular}{ccccccccc}
           \includegraphics[width=0.105\linewidth]
			{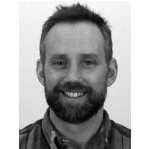}&
           \includegraphics[width=0.105\linewidth]
			{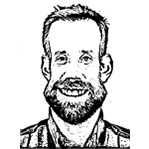}&
            \includegraphics[width=0.105\linewidth]
			{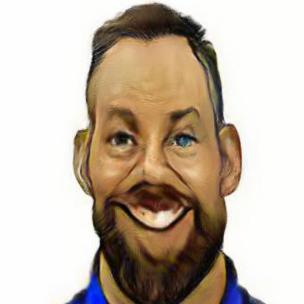} &
             \includegraphics[width=0.105\linewidth]
			{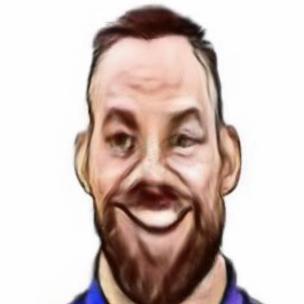} &
            \qquad   
            &
            \includegraphics[width=0.105\linewidth]
			{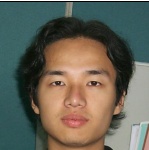}&
            \includegraphics[width=0.105\linewidth]
			{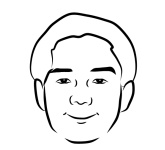}&
            \includegraphics[width=0.105\linewidth]
			{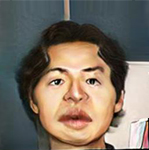}&
            \includegraphics[width=0.105\linewidth]
			{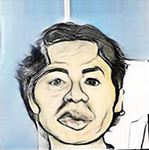}\\		
            Input & \citet{gooch2004human} & Ours 1&Ours 2&&
            Input & \citet{chen2002pictoon} & Ours 1&Ours 2\\
                      \includegraphics[width=0.105\linewidth]
			{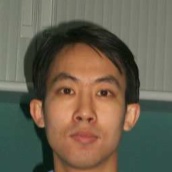}&
           \includegraphics[width=0.105\linewidth]
			{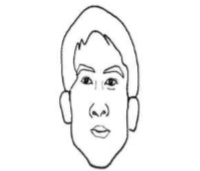}&
            \includegraphics[width=0.105\linewidth]
			{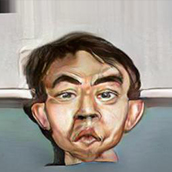} &
             \includegraphics[width=0.105\linewidth]
			{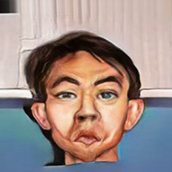} &
            \qquad   
            &
            \includegraphics[width=0.105\linewidth]
			{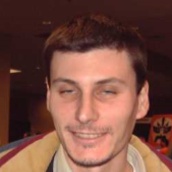}&
            \includegraphics[width=0.105\linewidth]
			{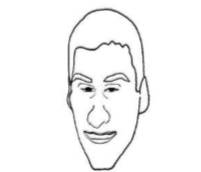}&
            \includegraphics[width=0.105\linewidth]
			{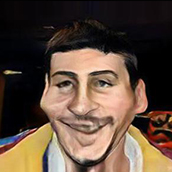}&
            \includegraphics[width=0.105\linewidth]
			{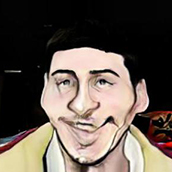}\\		
            Input & \citet{liang2002example} & Ours 1&Ours 2&&
            Input & \citet{liang2002example} & Ours 1&Ours 2\\
               \includegraphics[width=0.105\linewidth]
			{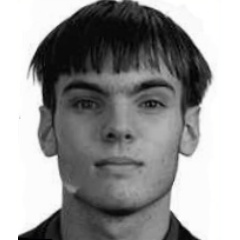}&
           \includegraphics[width=0.105\linewidth]
			{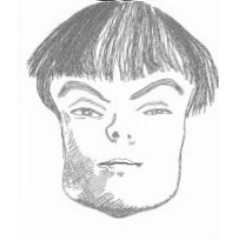}&
            \includegraphics[width=0.105\linewidth]
			{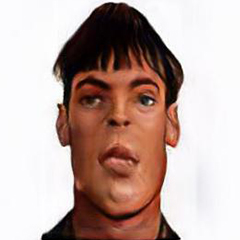} &
             \includegraphics[width=0.105\linewidth]
			{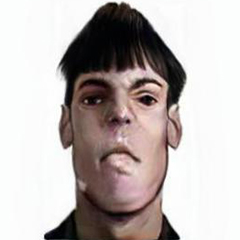} &
            \qquad   
            &
            \includegraphics[width=0.105\linewidth]
			{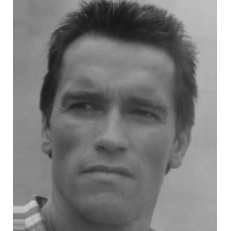}&
            \includegraphics[width=0.105\linewidth]
			{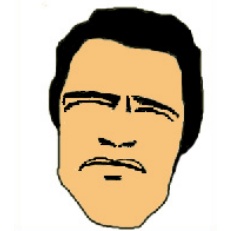}&
            \includegraphics[width=0.105\linewidth]
			{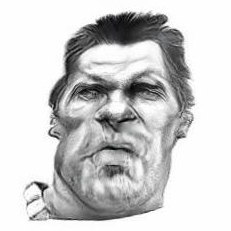}&
            \includegraphics[width=0.105\linewidth]
			{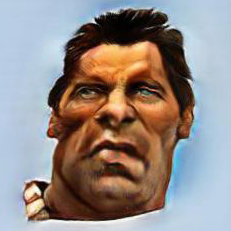}\\		
            Input & \citet{mo2004improved} & Ours 1&Ours 2&&
            Input & \citet{mo2004improved} & Ours 1&Ours 2\\      
		\end{tabular}}

    \caption{Comparison with graphics-based caricature generation techniques, including two interaction-based methods (\citet{gooch2004human}, \citet{chen2002pictoon}), one paired-example-based method (\citet{liang2002example}) and one rule-based method (\citet{mo2004improved}). Inputs are from their papers.}
	\label{fig:comp_tra}
\end{figure*}

\begin{figure*}
	\footnotesize
	\setlength{\tabcolsep}{0.002\linewidth}
    \scalebox{1.1}{
		\begin{tabular}{ccccccccc}
			\includegraphics[height=0.095\linewidth]
			{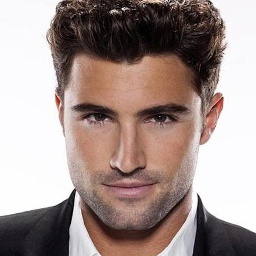}&
			\includegraphics[height=0.095\linewidth]
			{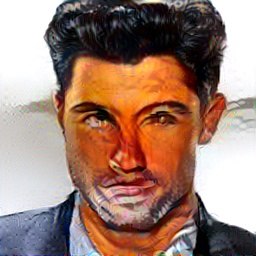}&
			\includegraphics[height=0.095\textwidth]
			{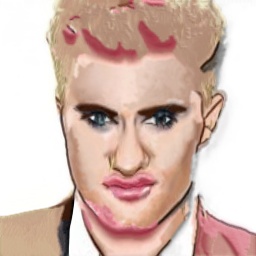}&
              \includegraphics[height=0.095\textwidth]
			{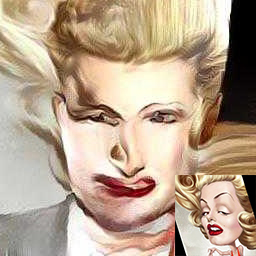}&
            \includegraphics[height=0.095\textwidth]
			{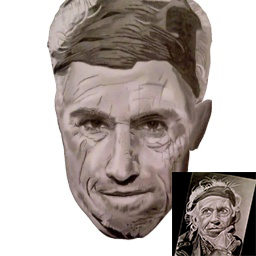}&
			\includegraphics[height=0.095\textwidth]
			{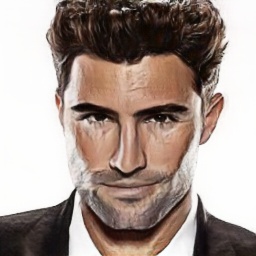}&
			\includegraphics[height=0.095\linewidth]
			{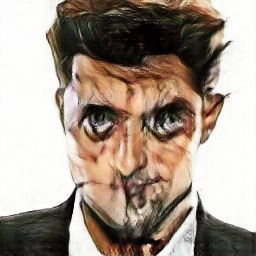}&
			\includegraphics[height=0.095\textwidth]
			{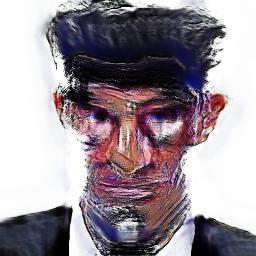}&      
			\includegraphics[height=0.095\textwidth]
			{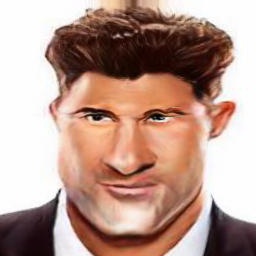}\\
      		\includegraphics[height=0.095\linewidth]
			{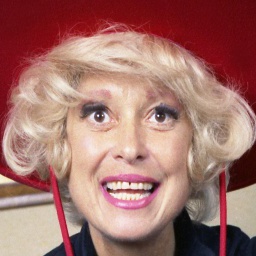}&
			\includegraphics[height=0.095\linewidth]
			{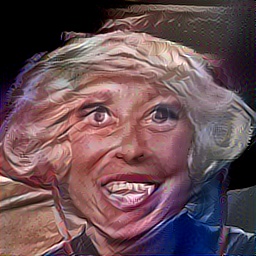}&
			\includegraphics[height=0.095\textwidth]
			{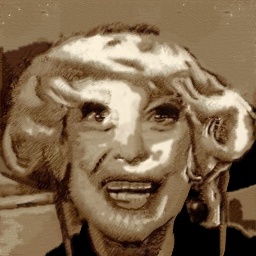}&
            \includegraphics[height=0.095\textwidth]
            {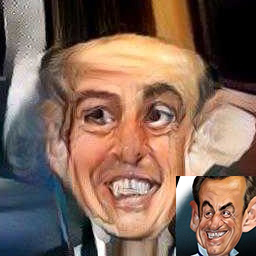}&
            \includegraphics[height=0.095\textwidth]
			{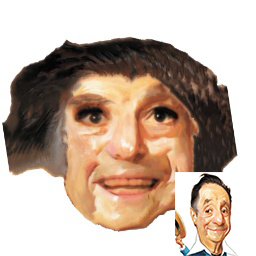}&
            \includegraphics[height=0.095\textwidth]
			{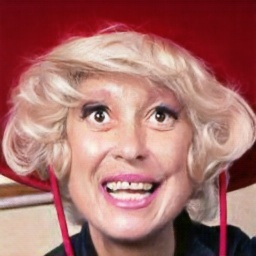}&
			\includegraphics[height=0.095\linewidth]
			{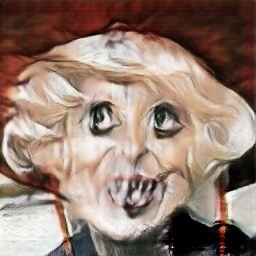}&
			\includegraphics[height=0.095\textwidth]
			{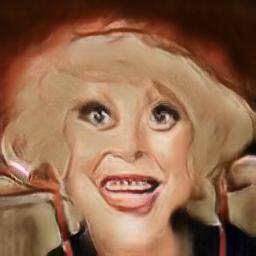}&     
			\includegraphics[height=0.095\textwidth]
			{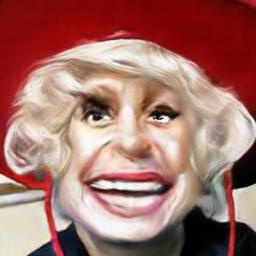}\\
            \includegraphics[height=0.095\linewidth]
			{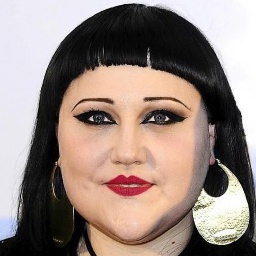}&
			\includegraphics[height=0.095\linewidth]
			{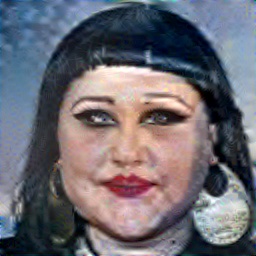}&
			\includegraphics[height=0.095\textwidth]
			{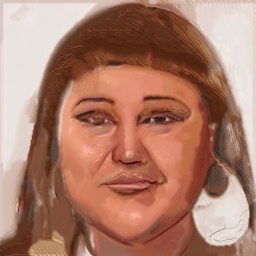}&
            \includegraphics[height=0.095\textwidth]
			{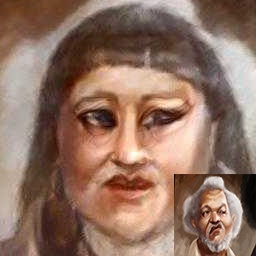}&
            \includegraphics[height=0.095\textwidth]
			{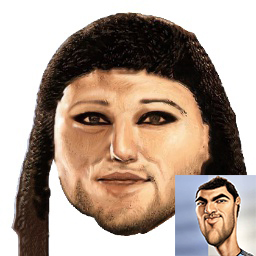}&
            \includegraphics[height=0.095\textwidth]
			{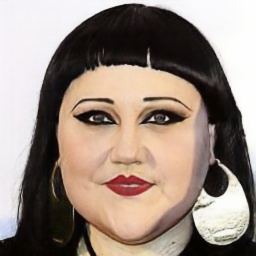}&
			\includegraphics[height=0.095\linewidth]
			{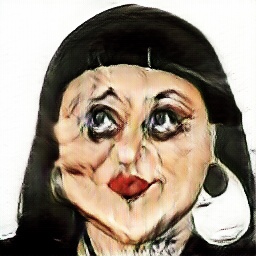}&
			\includegraphics[height=0.095\textwidth]
			{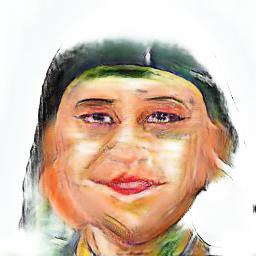}&      
			\includegraphics[height=0.095\textwidth]
			{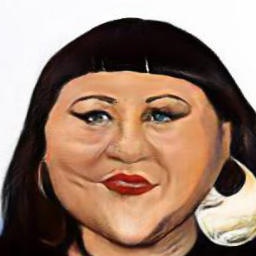}
		\\Input& Neural Style & Deep Analogy & Portrait Painting & Facial Animation&CycleGAN&UNIT&MUNIT&Ours
		\end{tabular}}

    \caption{Comparison with deep-learning-based methods, including two general image style transfer methods (neural style \citep{gatys2015neural} and Deep Analogy \citep{liao2017visual}), two face-specific style transfer methods (Portrait Painting \citep{selim2016painting} and Facial Animation \citep{fivser2017example}), two single-modal image translation networks (CycleGAN\citep{zhu2017toward}, UNIT \citep{liu2017unsupervised}) and one multi-modal image translation network (MUNIT\citep{huang2018munit}). Inputs are from CelebA dataset excluding the 10K images used in training. Input images: CelebA dataset.}
	\label{fig:comp_deep}
\end{figure*}

\subsection{Comparison to graphics-based methods}
We compare our \emph{CariGANs} with four representative graphics-based methods for caricature generation, including \citet{gooch2004human}, \citet{chen2002pictoon}, \citet{liang2002example} and \citet{mo2004improved}. We test on the cases from their papers and show the visual comparison in \fref{fig:comp_tra}. It can be seen that these compared methods focus less on the appearance stylization and only model some simple styles like sketch or cartoon, while ours can reproduce much richer styles by learning from thousands hand-drawn caricatures. As to the geometric exaggeration, \citet{gooch2004human} and \citet{chen2002pictoon} require manual specification. \citet{liang2002example} needs to learn the deformation form a pair of examples which is not easy to get in practice. \citet{mo2004improved} is automatic by exaggerating differences from the mean, but the hand-crafted rules are difficult to describe all geometric variations in the caricature domain. In contrast, our learning-based approach is more scalable.

\subsection{Comparison to deep-learning-based methods}
\label{sec:comp2}
We visually compare our CariGAN with existing deep-learning based methods in \fref{fig:comp_deep}. Here, all methods are based on author provided implementations with the default settings, except for \cite{selim2016painting} which has no code released and we implement ourselves. First we compare with style transfer techniques which migrate the style from a given reference. We consider two general style transfer methods (\citet{gatys2015neural} and \citet{liao2017visual}), and two methods tailed for faces (\citep{selim2016painting} and \citep{fivser2017example}). All the references used in these methods are randomly selected from our hand-drawn caricature dataset. As we can see, they can transfer the style appearance from the caricature to the input photo, but cannot transfer geometric exaggerations.

Our \emph{CariGANs} is compared with three general image-to-image translation networks, including two representative works (CycleGAN \citep{zhu2017toward} and UNIT \citep{liu2017unsupervised}) for single-modal unpaired image translation, and the sole network for multi-modal unpaired image translation (MUNIT \citep{huang2018munit}). We train these networks using the same dataset as ours. Since their networks should learn both two mappings of geometry and appearance jointly, this poses a challenge beyond their capabilities. UNIT and MUNIT fail to preserve the face structure. CycleGAN keeps the face structure but few artistic style and exaggeration learned. Thanks to the two GANs framework, our network better simulates hand-drawn caricatures in both geometry and appearance, while keeping the identity of the input.

\subsection{Perceptual study}
\begin{figure*}
\includegraphics[width=1.0\linewidth]{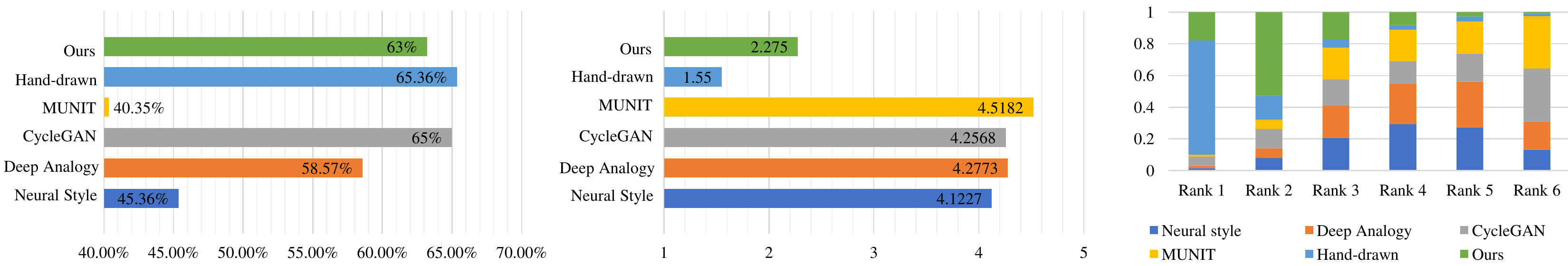}\\
	\footnotesize
	\setlength{\tabcolsep}{0.003\linewidth}
		\begin{tabular}{ccc}
			 \qquad  \qquad (a)   \qquad  \qquad  \qquad  &  \qquad  \qquad  \qquad  \qquad  \qquad   \qquad   \qquad   \qquad  (b)  \qquad \qquad  \qquad   \qquad  \qquad  \qquad  &   \qquad  \qquad  \qquad  \qquad  \qquad  \qquad  (c) 
		\end{tabular}
	\vspace*{-0.1in}
    \caption{User study results. (a) Percentages of correct face recognition in task 1. (b) Average rank of each method in task 2. (c) Percentage of each method that has been selected in each rank.}
	\label{fig:userstudy}
\end{figure*}

\begin{figure*}
	\footnotesize
	\setlength{\tabcolsep}{0.002\linewidth}
    \scalebox{1.09}{
		\begin{tabular}{ccccccc}
        	\includegraphics[height=0.125\linewidth]
			{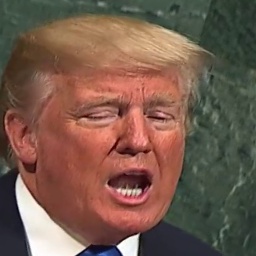}&
			\includegraphics[height=0.125\linewidth]
			{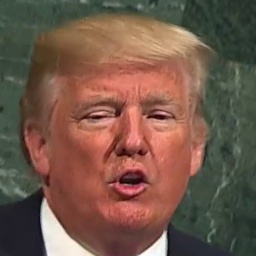}&
			\includegraphics[height=0.125\textwidth]
			{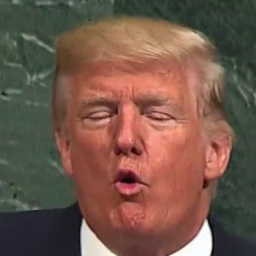}&
			\includegraphics[height=0.125\textwidth]
			{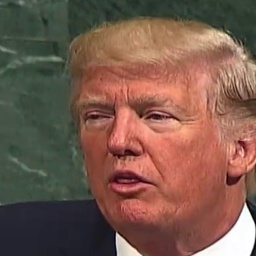}&
			\includegraphics[height=0.125\linewidth]
			{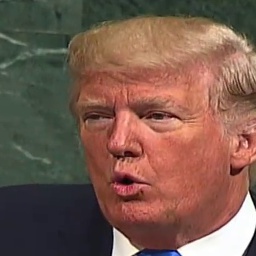}&
			\includegraphics[height=0.125\textwidth]
			{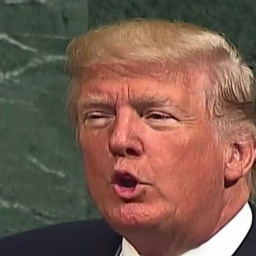}&
			\includegraphics[height=0.125\textwidth]
			{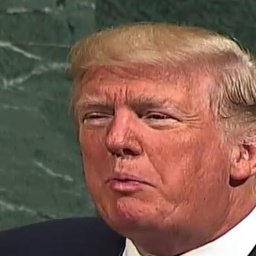}\\		
			\includegraphics[height=0.125\linewidth]
			{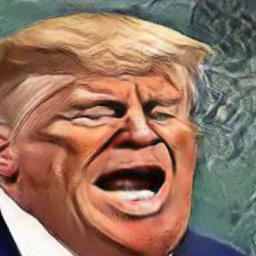}&
			\includegraphics[height=0.125\linewidth]
			{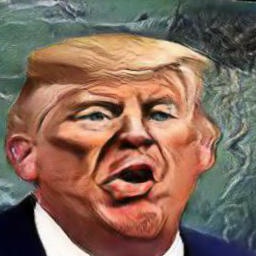}&
			\includegraphics[height=0.125\textwidth]
			{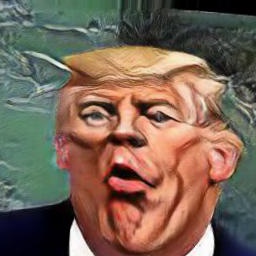}&
			\includegraphics[height=0.125\textwidth]
			{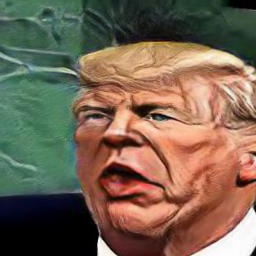}&
			\includegraphics[height=0.125\linewidth]
			{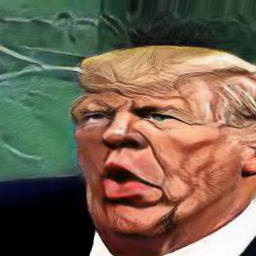}&
			\includegraphics[height=0.125\textwidth]
			{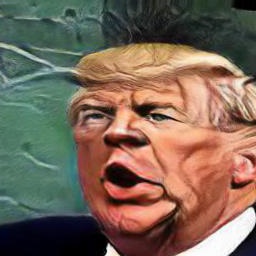}&
			\includegraphics[height=0.125\textwidth]
			{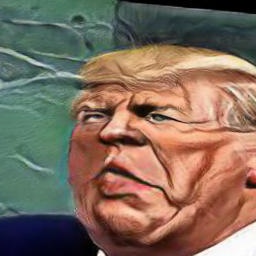}\\
		
		\end{tabular}}
	
    \caption{A video caricature example. The upper row shows the input video frames and the bottom row shows the generated caricature results. Trump video courtesy of the White House (public domain).}
	\label{fig:video}
\end{figure*}

We conduct two perceptual studies to evaluate our \emph{CariGAN} in terms of recognizability and style faithfulness. We compare caricatures drawn by artists with the results achieved by the following techniques: two neural style transfer methods (\citet{gatys2015neural},\citet{liao2017visual}), two image-to-image translation networks (CycleGAN \citep{zhu2017toward}, MUNIT \citep{huang2018munit}), and our \emph{CariGAN}. The compared methods are trained and tested in the same way as described in \Sref{sec:comp2}. 

We construct a new testing dataset with identity information for the two user studies. We randomly select 500 samples from our hand-drawn caricature dataset. For each caricature sample, we collect 20 different portrait photos of the same person from the Internet. All examples in the two studies are randomly sampled from this dataset, which are included in our supplemental material.

The first study assesses how well the identity is preserved in each technique. The study starts from showing 8 example caricature-photo pairs, to let the participant be familiar with the way of exaggeration and stylization in caricatures. Then 60 questions (10 for each method) follow. In each question, we present a caricature generated by our method or the method we compare it with, and ask the participant to select a portrait photo with the same identity as the caricature from 5 choices. Among the choices, one is the correct subject, while the other four items are photos of other subjects with similar attributes (\eg, sexual, age, glasses) to the correct subject. The attributes are automatically predicted by Azure Cognitive Service. Participants are given unlimited time to answer. We collect 28 responses for each question, and calculate the recognition rate for each method, shown in \fref{fig:userstudy} (a). 

As we can see, the results of MUNIT, Neural Style, and Deep Analogy pose more difficulty in recognizing as the correct subject, since the visual artifacts in their results mess up the facial features. We show examples in \fref{fig:comp_deep}. CycleGAN is good at preserving the identity because it is more conservative and produces photo-like results. Surprisingly, hand-drawn caricatures have the highest recognition rate, even better than the photo-like CycleGAN. We guess this is because professional artists are good at exaggerating the most distinct facial features which helps the recognition. Our recognition rate is also high, and very close to that of hand-drawn caricatures and photo-like results produced by CycleGAN. 

The second study assesses how close the generated caricatures are to the hand-drawn ones in visual styles. The study begins with the showcase of 8 caricatures drawn by artists, which lets the participant know what the desired caricature styles is. Later, we present one hand-drawn caricature, and five results generated by ours and compared methods, to participants in a random order, at every question. These 6 caricatures depict the same person. We ask participants to rank them from ``the most similar to given caricature samples" to ``the least similar to caricature". We use 20 different questions and collect 22 responses for each question.

As shown in \fref{fig:userstudy} (b), hand-drawn caricatures and ours rank as the top two. Our average rank is $2.275$ compared to their rank $1.55$. Other four methods have comparable average ranks but far behind ours. We further plot the percentages of each method that has been selected in each rank (\fref{fig:userstudy} (c)). Note that ours is ranked better than the hand-drawn one $22.95\%$ of the times, which means our results sometime can fool users into thinking it is the real hand-drawn caricature. Although it is still far from an ideal fooling rate (\ie, $50\%$), our work has made a big step approaching caricatures drawn by artists, compared to other methods.

\section{Extensions}
We extend of CariGANs to two interesting applications.

\subsection{Video caricature}
We directly apply our CariGANs to the video frame by frame. Since our CariGANs exaggerate and stylize the face according to facial features. The results are overall stable in different frames, as shown in \fref{fig:video}. Some small flickering can be resolved by adding temporal constraint in our networks, which is left for future work. The video demo can be found in our supplemental material.

\subsection{Caricature-to-photo translation}
\begin{figure}
	\footnotesize
	\setlength{\tabcolsep}{0.003\linewidth}
    \scalebox{1.1}{
		\begin{tabular}{cccccccc}
			\includegraphics[height=0.21\linewidth]
			{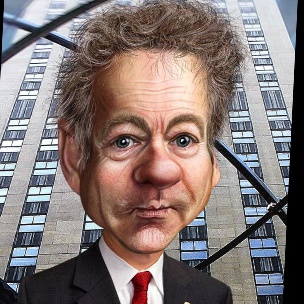}&
            \includegraphics[height=0.21\linewidth]
			{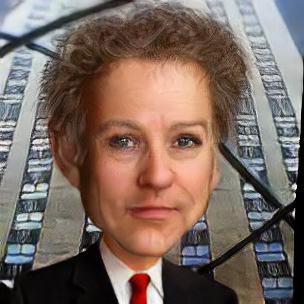}&
            \includegraphics[height=0.21\linewidth]
			{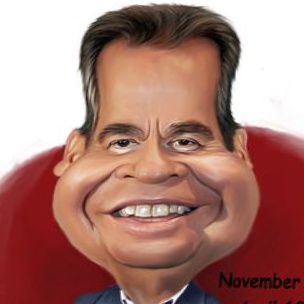}&
            \includegraphics[height=0.21\linewidth]
			{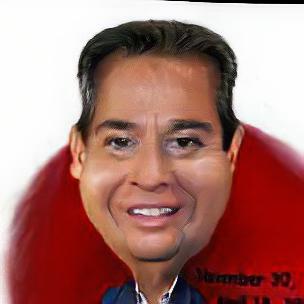}
            \\
            \includegraphics[height=0.21\linewidth]
			{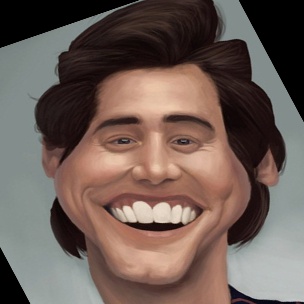}&
            \includegraphics[height=0.21\linewidth]
			{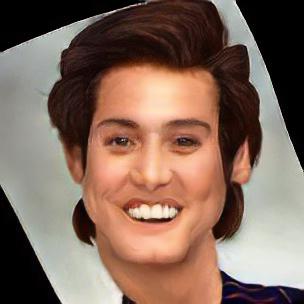}&
            \includegraphics[height=0.21\linewidth]
			{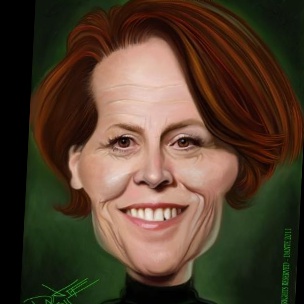}&
            \includegraphics[height=0.21\linewidth]
			{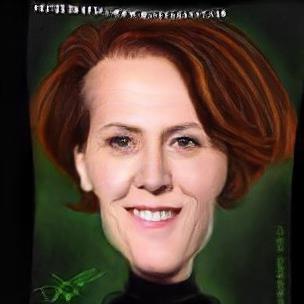}\\
            Input & Output & Input & output 
		\end{tabular}}
	\vspace{-1em}
    \caption{Converting caricatures to photos. Inputs (from left to right, top to bottom): \textcopyright DonkeyHotey/Flickr, \textcopyright Rockey Sawyer/deviantart, \textcopyright Guillermo Ram\'{i}rez/deviantart, \textcopyright Michael Dante/wittygraphy.} \vspace{-1em}
	\label{fig:tophoto}
\end{figure}

Since both \emph{CariGeoGAN} and \emph{CariGeoStyGAN} are trained to learn the forward and backward mapping symmetrically, we can reverse the pipeline (\fref{fig:Pipeline}) to convert an input caricature into its corresponding photo. Some examples are shown in \fref{fig:tophoto}. We believe it might be useful for face recognition in caricatures. 

\begin{figure}
	\footnotesize
	\setlength{\tabcolsep}{0.003\linewidth}
    \scalebox{1.1}{
		\begin{tabular}{ccc}
			\includegraphics[height=0.27\linewidth]
			{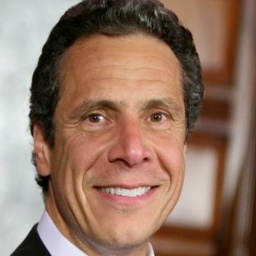}&
            \includegraphics[height=0.27\linewidth]
			{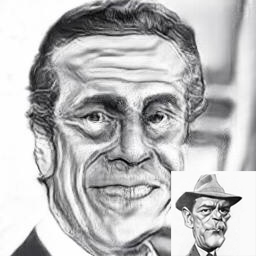}&
            \includegraphics[height=0.27\linewidth]
			{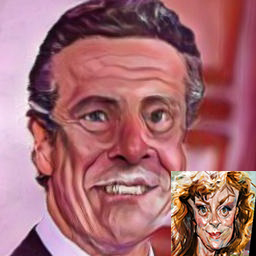}\\
          
            (a) Input & (b) Result (ref1) & (c) Result (ref2)
		\end{tabular}}
		
    \caption{Our result is faithful to the reference style which is common in the caricature dataset (b), but is less faithful to some uncommon style (c). Input image: CelebA dataset.}	
    \vspace{-0.2in}
	\label{fig:limit}
\end{figure}

\section{Conclusions}
We have presented the first deep learning approach for unpaired photo-to-caricature translation. Our approach reproduces the art of caricature by learning both geometric exaggeration and appearance stylization respectively with two GANs. Our method advances the existing methods a bit in terms of visual quality and preserving identity. It better simulates the hand-drawn caricatures to some extent. Moreover, our approach supports flexible controls for user to change results in both shape exaggeration and appearance style.  

Our approach still suffers from some limitations. First,  Our geometric exaggeration is more obviously observed in the face shape than other facial features and some small geometric exaggerations on ears, hairs, wrinkles and etc., cannot be covered. That is because there are 33 out of total 63 landmarks lying on the face contour. Variants of these landmarks dominate the PCA representation. This limitation can be solved by adding more landmarks. Second, it is better to make our \emph{CariGeoGAN} to be multi-modal as well as our \emph{CariStyGAN}, but we fail to disentangle content and style in geometry since their definitions are still unclear. As to the appearance stylization, our results are faithful to the reference style which are common in the caricature dataset (\eg, sketch, cartoon) but are less faithful to some uncommon styles (\eg, oil painting), as shown in \Fref{fig:limit}. That is because the our \emph{CariStyGAN} cannot learn the correct style decoupling with limited data. Finally, our \emph{CariStyGAN} is trained and tested with low-res ($256 \times 256$) images, we consider applying the progressive growing idea from \citep{karras2017progressive} in our \emph{CariStyGAN} to gradually add details for high-res images (\eg, 1080p HD). These are interesting, and will explored in future work.     

\begin{acks}
We want to thank the anonymous referees for their valuable comments and  helpful suggestions. We also want to thank the participants in our user study, the artists for allowing us to use their works, and authors of \cite{fivser2017example} for helping us generate the comparison examples with their method.
\end{acks}

\bibliographystyle{ACM-Reference-Format}
\bibliography{carigan}
\end{document}